\documentclass[conference]{IEEEtran} 
\usepackage{cite}
\usepackage{amsmath,amssymb,amsfonts}
\usepackage{graphicx}
\usepackage{textcomp}

\usepackage{multicol, multirow}
\usepackage[ruled,vlined]{algorithm2e}
\usepackage{xcolor}
\usepackage{lipsum}
\usepackage{times}
\usepackage{rotating}
\usepackage{gensymb}
\usepackage{breakurl}
\usepackage{subfigure}

\usepackage{url}
\usepackage[none]{hyphenat}
\usepackage{comment}
\usepackage{authblk}
\def\BibTeX{{\rm B\kern-.05em{\sc i\kern-.025em b}\kern-.08em
    T\kern-.1667em\lower.7ex\hbox{E}\kern-.125emX}}

\providecommand{\keywords}[1]
{
  \small	
  \textbf{\textit{Keywords---}} #1
}
    
\begin{document}
%\history{Date of publication xxxx 00, 0000, date of current version xxxx 00, 0000.}
%\doi{10.1109/ACCESS.2017.DOI}

\title{Weed Density and Distribution Estimation for Precision Agriculture using Semi-Supervised Learning }
\author[1]{Shantam Shorewala}
\author[*1]{Armaan Ashfaque}
\author[*1]{Sidharth R }
\author[$\dag$ 1]{Ujjwal Verma \footnote{Corresponding Author: ujjwal.verma@manipal.edu}}
\affil[1]{ Manipal Institute of Technology, Manipal Academy of Higher Education, Manipal, India. }
\affil[$\dag$]{Corresponding author: ujjwal.verma@manipal.edu}
%\affil[2]{Department of Information \& Communication Engineering, Manipal Institute of Technology, Manipal Academy of Higher Education, Manipal, India. }
%\affil[3]{Department of Computer Science Engineering, Manipal Institute of Technology, Manipal Academy of Higher Education, Manipal, India. }

% \tfootnote{This paragraph of the first footnote will contain support 
% information, including sponsor and financial support acknowledgment. For 
% example, ``This work was supported in part by the U.S. Department of 
% Commerce under Grant BS123456.''}

% \markboth
% {Author \headeretal: Preparation of Papers for IEEE TRANSACTIONS and JOURNALS}
% {Author \headeretal: Preparation of Papers for IEEE TRANSACTIONS and JOURNALS}

%\corresp{Corresponding author: Ujjwal Verma (e-mail: ujjwal.verma@manipal.edu).}

\maketitle
\thispagestyle{plain}
\pagestyle{plain}
\begin{abstract}
Uncontrolled growth of weeds can severely affect the crop yield and quality. Unrestricted use of herbicide for weed removal alters biodiversity and cause environmental pollution. Instead, identifying weed-infested regions can aid selective chemical treatment of these regions. Advances in analyzing farm images have resulted in solutions to identify weed plants. However, a majority of these approaches are based on supervised learning methods which requires huge amount of manually annotated images. As a result, these supervised approaches are economically infeasible for the individual farmer because of the wide variety of plant species being cultivated. In this paper, we propose a deep learning-based semi-supervised approach for robust estimation of weed density and distribution across farmlands using only limited color images acquired from autonomous robots. This weed density and distribution can be useful in a site-specific weed management system for selective treatment of infected areas using autonomous robots. In this work, the foreground vegetation pixels containing crops and weeds are first identified using a Convolutional Neural Network (CNN) based unsupervised segmentation. Subsequently, the weed infected regions are identified using a fine-tuned CNN, eliminating the need for designing hand-crafted features.  The approach is validated on two datasets of different crop/weed species (1) Crop Weed Field Image Dataset (CWFID), which consists of carrot plant images and the (2) Sugar Beets dataset. The proposed method is able to localize weed-infested regions a maximum recall of 0.99 and estimate weed density with a maximum accuracy of 82.13\%. Hence, the proposed approach is shown to generalize to different plant species without the need for extensive labeled data. 
\end{abstract}


%%%%%%%%%%%%%%%%%%%%%%%%
\section{Introduction}
\label{introduction}

Agriculture continues to be the most important industry across the world necessary for sustaining mankind. There have been significant improvements in the machinery operated by the farmers to cultivate their lands. One common aspect of farming is \textit{weeding} - it refers to the removal or treatment of weed plants. Weeds are undesirable plants that compete with crop plants for natural resources such as sunlight, minerals, and water. Hence it becomes necessary to selectively remove these plants to ensure a healthy crop yield \cite{12,wang_overview}.
However, the traditional practice of treating the entire farmland indiscriminately with agrochemicals for weed control, in addition to being expensive, adversely impacts the soil biodiversity, quality of freshwater available to humans as well as the human health \cite{soil_health_earthworms,impact_2,water,human_health}. An alternative to chemical weeding is to manually pick the weed plants (manual weeding). This approach, however, is time and labour intensive. 

Precision Agriculture is a ``management strategy that takes account of temporal and spatial variability to improve the sustainability of agricultural production" \cite{ispa_def}. Common applications of precision agriculture include identification of weeds, crop and soil health monitoring, site-specific management for tasks such as tillage, sawing, mechanical weeding and distribution of fertilizers, crop yield estimation, fruit/vegetable detection and picking \cite{pa_overview, pa_2, pa_3, pa_example, Ujjwal-Seg1, Ujjwal-Seg2, Ujjwal_Seg3}.

Autonomous robots have been used for chemically weeding patches of weed plants \cite{aut_robot_1, aut_robot_2, aut_robot_3}. These robots rely on systems, including machine vision, to identify and localize weed plants.  A typical image processing based weed detection approach consists of four stages: pre-processing, segmentation, feature extraction, and classification. The pre-processing prepares the input image for segmentation and typically consists of various image enhancement methods such as color space transformation. Subsequently, the enhanced image is segmented into two regions: vegetation and background. The segmentation procedure can be grouped into two categories: Index-based and learning based. The index-based approach differentiates between vegetation and background by comparing each pixel's intensity value with a threshold parameter. This approach is usually not robust to varying lighting conditions and overlapping crop and weed plants \cite{seg-1, seg-2, seg-3}. The learning-based methods for vegetation segmentation have been shown to overcome this challenge and are the preferred approach to accurately identify the vegetation \cite{wang_overview}. The segmentation procedure produces the vegetation mask that contains both crop and weed pixels in the same class. Therefore, a hand-crafted feature vector is computed based on biological morphology, spectral features, visual textures, and spatial contexts of the crop and weed plants. These feature vectors are then fed to a classifier to identify weeds from the segmented vegetation mask. The image processing based approach overcomes labour and time-intensive demands of manual weeding in addition to reducing the amount of chemical sprayed. However, the use of hand crafted features restricts the usage of these approaches to a particular crop/weed species. Recently, deep learning based approaches \mbox{\cite{1, 2, 3}} have been proposed which eliminates the need for hand crafted features. However, majority of these approaches are supervised approaches which require a huge amount of training data, thus limiting its application to few crop/weeds. The major challenges to a reliable, scalable vision system for the autonomous robots are (1) varying lighting conditions, (2) overlapping and occluded weed and crop plants, (3) varying weed density, and (4) different species of crop and weed plants. In addition, the supervised learning-based approach depends on the availability of annotated data.    It may be noted that there exist other sensors such as visible and near-infrared (Vis-NIR) spectroscopy, LiDAR, and sonar \cite{wang_overview,crop_lidar, crop_sonar} for weed identification. However, this study focuses on an image-based system for weed identification.

There also exists image classification based approaches for weed detection. In this approach, the entire image is labelled as a particular weed species, based on the weed species present in the image \mbox{\cite{deepweed}}. This approach is able to identify the weed present in the field but would not be able to compute the weed density. In this work, we propose an alternative patch based approach, which eliminates the need for pixel wise annotation, and can compute weed density and distribution. The primary objective of our work is to evaluate a semi-supervised pipeline for weed localization and density estimation in order to minimize the amount of manually annotated data required to train the deep networks. By reducing the dependence on data-intensive segmentation networks, we can enhance the adoption rate for different species of crops/weeds and in different environments/settings.

% \cite{26} describes a crop/weed identification system in line with this approach. However, they use multi-spectrum images (NIR, NDVI, and RGB) to segment the vegetation pixels. 

The main contribution of our work is a semi-supervised decision support system for robust estimation of weed distribution and density from a single color image acquired using an autonomous robot. Instead of focusing on pixel-wise segmentation, we seek to address the more fundamental question of which regions should be selectively treated with agrochemicals. This decision can be on the basis of estimated \textit{weed distribution or localization} and \textit{weed density}. The proposed approach can 

\begin{itemize}
    \item Robustly identify weed infected regions
    \item Compute weed density in the infected regions
    \item Enhance scalability and generalizability as it does not require pixel-wise annotations unlike end-to-end deep learning segmentation networks
\end{itemize}

The proposed approach leverages unsupervised Convolutional Neural Network to cluster the pixels into vegetation and background class. It is worth noting that any foreign objects or non-soil, non-vegetation pixels are also classified as background in the proposed approach. The vegetation mask is then overlaid on the input color image which is divided into smaller tiles. Each tile with vegetation coverage is then passed through a classifier that labels it either as weed or crop. Algorithm \ref{algo:overview} briefly describes the different stages in the proposed approach. Unlike existing image-based methods for weed classification, the proposed approach does not rely upon hand-crafted features. Moreover, the proposed approach does not require extensive segmentation labeling of crop and weed plant pixels as used in \cite{1,2,3,6}.

\begin{algorithm}
\SetAlgoLined
    \textbf{Input}: Color image ($I_{RGB}$) of the field acquired from an autonomous robot; \\
    \textbf{Output}: Weed density and distribution;\\
    Given ($I_{RGB}$), Generate the vegetation mask ($I_{veg}$) using CNN based unsupervised segmentation; \\
    Overlay $I_{RGB}$ with $I_{veg}$ to get $I_{masked}$; \\
    Divide the image $I_{masked}$  into smaller regions $I_{tile}$ (square tiles);\\
    \For{($I_{tile}$ in $I_{masked}$)}
    {
        Classify $I_{tile}$ into crop, weed or background;\\ \If{$I_{tile}$ is weed}{Estimate weed density}
    }
    \caption{Weed distribution and density estimation}
    \label{algo:overview}
\end{algorithm}

% without impacting the ability to estimate weed density. 
%Besides, the formulation of a two cluster unsupervised segmentation task also eliminates the need for determining the optimal number of clusters for an unsupervised weed segmentation, which can be a challenging task \cite{12}. 

The rest of the paper is structured as follows: Section \ref{related_work} discusses existing approaches to identify weed plants. Section \ref{methodology} describes individual steps of the proposed approach in detail. Datasets used and results are discussed in the following section. Finally, the conclusions drawn are presented in Section \ref{conclusion}.

%%%%%%%%%%%%%%%%%%%%%%%%%%%%%%%%%%%%%%%%%%%%%%%%%%%%%%%%

\section{Related work}
\label{related_work}

This section summarizes existing traditional as well as deep learning-based approaches for image-based weed detection and classification. For a detailed discussion, reader is referred to \cite{wang_overview}. Recent advances in the field of deep learning have been applied to precision agriculture to improve the limitations and inflexibility of traditional methods. A review of state-of-the-art deep learning approaches to disparate problems in agriculture, including identification of weeds, land cover classification, and fruit counting, among others, can be found in \cite{dl_survey}. 

% \paragraph{Weed detection using image processing} Initial image processing approaches, relying on index-based segmentation, extract blobs of green pixels i.e., vegetation blobs. This approach is usually not robust to varying lighting conditions and overlapping crop and weed plants (\cite{seg-1, seg-2, seg-3}). There have been efforts to address the limitations of index-based methods (\cite{robust_index_based}). Despite the improvement in robustness, the approach is limited only to binary segmentation (background and plant pixels). \cite{26} describe a crop/weed identification system as per the approach described in Section \ref{introduction}. However, they use multi-spectrum images (NIR, NDVI, and RGB) to segment the vegetation pixels and hand-crafted features to train a random forest classifier to discriminate between crop and weed plants. 

\paragraph{Supervised Learning} In the last few years, deep learning methods have achieved state-of-the-art results on challenging datasets for various applications such as autonomous driving \cite{17, 27}. However, they are generic, designed to handle a large number of object classes. For weed identification and mapping, a much smaller number of classes need to be handled. Multiple research works have previously proposed an end-to-end semantic segmentation network, built upon earlier works such as SegNet \cite{17}, that distinguish between crop and weed plants \cite{1, 2, 3}. In \cite{1}, networks is trained on 465 multispectral images and achieves extremely high F1 scores (>0.95). While the number of training images is relatively small, it does rely on images captured from multispectral sensors which results in higher costs. Authors of \cite{2,3} obtained comparable performance (F1 score > 0.90) with networks trained on a set of 10,000 RGB images. These results establish the feasibility of training deep learning models to discriminate crop and weed plants.  However, as with all supervised learning models, they require an extensively manually annotated dataset to train the network. This challenge is not as prominent in applications where models can generalize reasonably well to different settings without loss of performance (such as object detection for common items such as chairs, humans, etc).  Authors of \cite{2} also study the adaptability of their work to different plants by testing the trained network on a different set, achieving accurate results and demonstrating the need for adaptable networks. However, the datasets compared are similar in terms of background visual features of the vegetation. Instead, in our study, we propose an alternative approach that to pixel-wise segmentation models. 
% They extend their work \cite{6} to include another decoder to identify and localize plant stem to aid the separation of crop and weed plants. \
In \cite{rasti2019supervised}, authors utilise scatter transforms to produce feature vectors for an SVM to classify culture crops. The approach is trained on a synthetic dataset and achieves an accuracy of around 85\% on culture crops. Another approach to supervised learning that has been used can be found in \cite{RAJA2019278}. This approach uses synthetic markers for crops that are planted to accurately detect them via computer vision and achieve very high results of around 99.7\%. On the other hand, our approach takes as input raw RGB images with no types of augmentations or physical markers placed on the field.

An object detection based approach is proposed in \cite{25} for weed identification. A deep neural network is trained to produce coverage maps and bounding boxes for localization of crops and weeds. While achieving accurate results, this is a very data-intensive approach that requires both covering maps and bounding boxes to be manually annotated. In a separate study \cite{4},  multispectral orthomosaic maps are generated by projecting a 3D point cloud onto the ground plane. They propose to overcome the challenge of scanning a large area while preserving the fine details of plant distribution. A modified SegNet model is then used to segment the weed pixels in these maps. Such an approach is data-intensive (the study used a dataset with more than 10,000 images), requiring sensors that can produce point clouds besides having to train an end-to-end segmentation model. Another study by \cite{10} proposes a two-stage network that uses an end-to-end segmentation network to first create a binary vegetation mask. Vegetation blobs are then passed as patches to a deep VGG-16 network for classification. The two-stage pipeline is an useful technique but both the networks require training on the chosen types of crop fields. Our study builds upon the idea of a two-stages for identifying weed infestation by leveraging unsupervised learning for vegetation segmentation (which is the first stage). This leaves only tile labels to be generated for training the classifier. Thus, the use of these modules aids in reducing the data dependency and can be easily extended to any crop/weed combination.  

\paragraph{Transfer learning} The authors in \cite{5} proposed a weed classifier which utilizes features extracted from a pre-trained sparse autoencoder \cite{autoencoder_1, autoencoder_2}. However, the algorithm makes two simplifications. Firstly, the example patches from the aerial images used are pre-selected hence making the pipeline semi-automated. Secondly, the dataset being used is balanced, which, in reality, is not the case with crop-weed datasets. Previous works such as \cite{13} have tried to address the dependency on manually annotated extensive datasets. Using a pre-trained network \cite{29}, the authors trained a much smaller network compared to others. The results show that the network is able to generalize well to a small dataset without compromising on performance, achieving a best of 93.9\% accuracy.

\paragraph{Semi-supervised and unsupervised learning} Semi-supervised  and unsupervised learning methods have also been studied to perform weed detection. For instance, a comparative study of two deep unsupervised learning algorithms JULE \cite{21} and DeepCluster \cite{22}  is presented by \cite{12}, along with a deep network like VGG-16 \cite{19} or ResNet-50 \cite{14} to help classify and automatically label different classes of weeds. \cite{9} use K-Means pre-training to adjust network weights before a LeNet-5 \cite{24} model is used to classify the type of weeds. These approaches do not predict a dense map for weed or weed pixels, only the class to which the image belongs. Hence, they cannot estimate the weed density, which is imperative as variable spraying of herbicide leads to increased application efficiency and reduced environmental impact   \cite{rocha2015weed}\cite{shiratsuchi2003aplicaccao}\cite{fontes2005levantamento}. Authors of \cite{unsupervised_weed_scouting} propose an unsupervised approach to cluster plants into different classes. They achieve competitive results under the assumption that none of the plants (either weed or crop) overlap each other. In practice, it is not an assumption that will hold true for varying plant species. Moreover, one of the challenging tasks in the unsupervised approach is determining the optimal number of clusters in which the image should be segmented \cite{12}. In the proposed work, we alleviate this difficulty by utilizing the unsupervised approach for only segmenting the vegetation mask, thereby fixing the number of the cluster as two. 

% Mordifferent cleover, one o into f the challenging tasks in the unsupervised approach is determining the optimal number of clusters in which the image should be segmented (\cite{12}). In the proposed work, we alleviate this difficulty by utilizing the unsupervised approach for determining vegetation mask, thereby fixing the number of the cluster as two. 

The method described by \cite{10} is closest to the proposed approach. The authors utilize a deep learning-based method for weed identification. A two-stage network was used: first, CNN extracts the vegetation mask while the second CNN identifies weeds from crops. However, there is a significant difference when it comes to the components used in the proposed work. Compared to supervised learning networks adopted by \cite{10}, the proposed method requires significantly less training data (vegetation segmentation is unsupervised while the classifier is trained with a small number of region labels). The dataset used in \cite{10} consisted of 2000 images while the proposed approach is tested on network trained on datasets with 90 and 500 images respectively (including augmented images); this highlights the significant reduction in the number of images. This contrast is further increased by the type of annotation required - proposed pipeline eliminates the need for pixel-wise annotations whereas the networks in \cite{10} need pixel-wise annotated images for training. Only the classifier needs to be fine-tuned with images of new plant species and binary labels in the proposed work. Further, they mention that a large percentage of errors arise due to overlapping plants. The proposed approach is shown to be both robust to poor illumination, occlusions, and plant density as well as adaptable to varying plant species. Unlike other semi-supervised approaches, the proposed method can still robustly estimate both the weed density and distribution, from RGB images.

\paragraph{Weed Density Estimation:} Weed density is an important parameter which helps in identifying the regions to be treated with chemicals\cite{rocha2015weed}\cite{shiratsuchi2003aplicaccao}\cite{fontes2005levantamento}. In \cite{wd-1, wd-2}, authors propose methods to estimate weed densities in row crops. \cite{wd-1} describes the weed density with cluster rate (ratio of weed quantity to land area) and weed pressure (ratio of weed quantity to crop) parameters. \cite{wd-2} extract the weed distribution using a positional histogram. The histogram is plotted by counting the number of white pixels in a binary vegetation mask along each column to obtain the lateral pixel distribution. Weed density (ratio of weed pixels in each interval to the image size) is obtained for fixed intervals. This approach is suitable only for only inter-row weed plants and does not account for weed-crop overlapping. Also, prior knowledge of crop row positions to estimate weed densities is assumed. Hence, it is not easily adaptable to different kinds of crop plantations. 

% In comparison, the proposed approach can handle overlapping crop/weed plants and does not assume prior information about crop positioning. Thus, it can better describe weed density and distribution for a variety of crop/weed plants.

% The rest of the paper is structured as follows: Section \ref{m&m} describes the individual steps in the proposed approach in detail, followed by a description of the datasets used. The results are highlighted and discussed in the following section. Finally, the conclusions drawn are presented in Section \ref{section:conclusion}.

\begin{figure*}[t!]
    \includegraphics[width=\textwidth, height=0.45\textwidth]{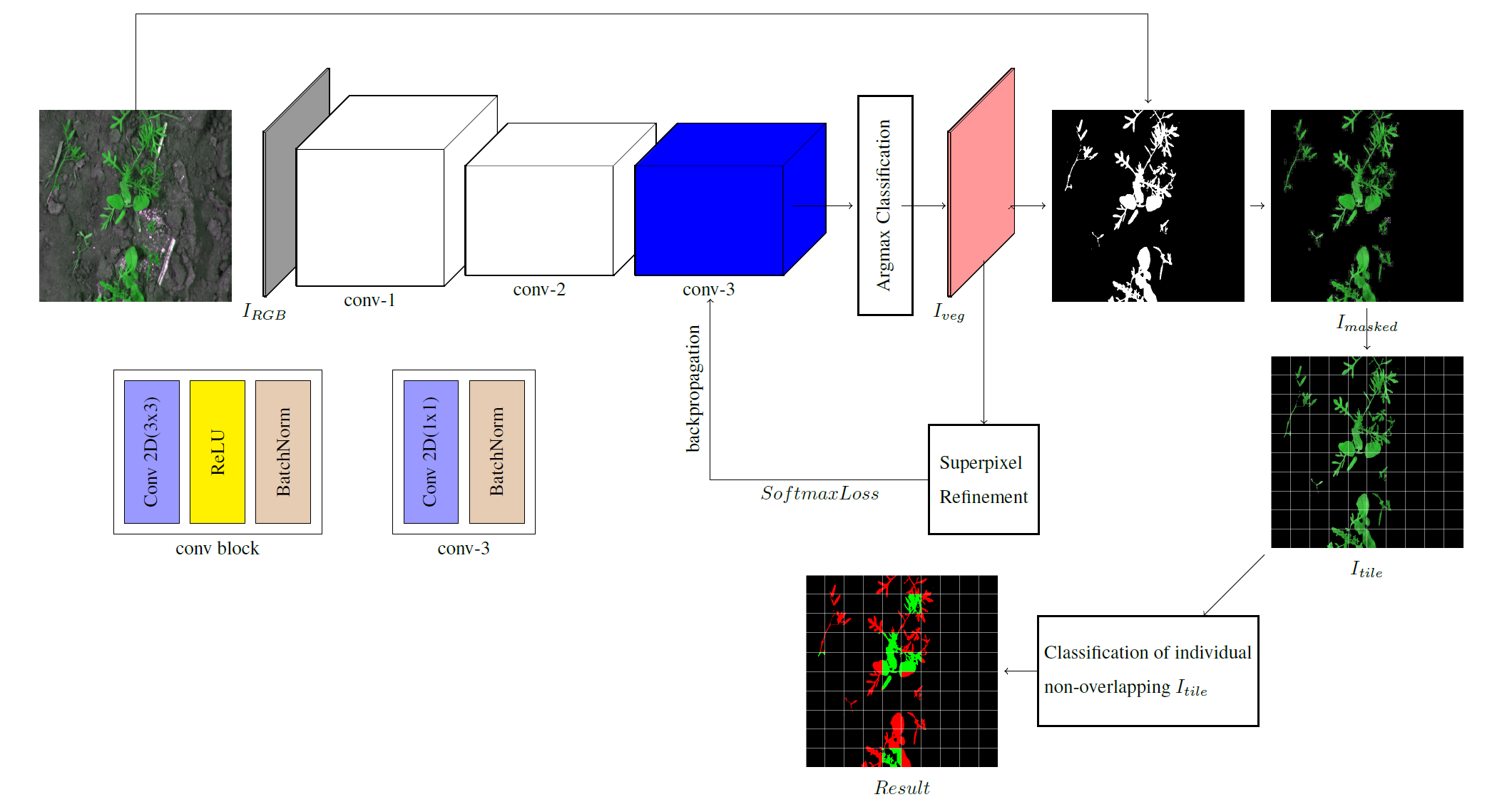}
    \caption{Overview of the proposed approach: An unsupervised CNN based binary segmentation is applied to the input color image to generate the vegetation mask. Subsequently, the masked image is sub-divided into non-overlapping tiles, which are then passed through a classifier. This classifier classifies the tiles as weed, crop, or background. }
    \label{fig:overview}
\end{figure*}

%%%%%%%%%%%%%%%%%%%%%%%%%%%%%%%%%%%%%%%%%

\section{Methodology}
\label{methodology}

In order to selectively treat the farmland under cultivation, the proposed approach identifies the weed-infested regions and the corresponding weed density. A single RGB image is taken as the input for the pipeline. Image pixels are first clustered into two classes (vegetation and background) using an unsupervised deep learning-based segmentation network. This process generates a vegetation mask (foreground) and a background mask. The vegetation mask, $I_{veg}$, is then overlaid on the original RGB image to get the region of interest (RoI) denoted by $I_{masked}$. This is then divided into smaller regions or patches $I_{tile}$ (square tiles). For each tile ($I_{tile}$), a corresponding feature vector is extracted that describes the properties of the vegetation pixels present in the tile. These vectors are then used to classify $I_{tile}$ as either crop or weed plant using a binary classifier. In addition, the performance of a fine-tuned CNN (ResNet50) is also studied for classifying $I_{tile}$. The location of weed-infested regions can be inferred from the regions $I_{tile}$, which are classified as weed label. Weed density in the region can be estimated from the vegetation pixel density in the area. The ratio of the number of vegetation pixels in each region, classified as either crop or weed, to the region's total land area in pixels gives the corresponding density estimate. It may be noted that only a part of the proposed method is trained in a supervised manner, resulting in a scalable approach that can be adapted for different weed and crop plant species. Figure \ref{fig:overview} provides an overview of the proposed pipeline. The rest of the section describes the individual steps in detail.  
% as well as information about the dataset used. 

\subsection{Vegetation segmentation}
\label{method_segmentation}

The input image ($I_{RGB}$) is first resized to 500x500 sq. pixels using the bicubic interpolation method implemented by OpenCV library \cite{opencv}. Each pixel in the image has to be clustered into one of the two classes - background or vegetation. For this purpose, we use the CNN based approach proposed in \cite{unsupervised_net} for unsupervised segmentation. However, to make the paper self-contained, the work is described in brief. This iterative approach is solved in two steps: label prediction assuming fixed network parameter (forward pass of the network) and learning network parameters through back-propagation assuming fixed (predicted) labels.  The approach proposes the following constraints for predicting the cluster or class to which each pixel might belong - (1) The first constraint is on feature similarity. Pixels that are similar to each other are clustered together. In order to achieve this, a response map for all the pixels is generated. Based on it, each pixel is assigned to a cluster to which it is closest according to the response map, (2) The second constraint is on spatial continuity. The authors use a number of superpixels extracted from the image and force the same cluster label for all the labels within. Superpixels can be defined as a group or cluster of pixels that exhibit common characteristics such as pixel intensity and proximity. The network extracts the superpixels using the Simple Linear Iterative Clustering (SLIC) algorithm \cite{slic} which operates in a five-dimensional space (three channels of CieLab colorspace and 2D image coordiantes (x, y)), (3) Final constraint is placed on the number of unique clusters into which the image is segmented. Given a maximum number of clusters $q$, the preference is for a large number of classes to avoid under segmentation. The solution for this constraint is to perform intra-axis normalization on the response map before assigning the cluster labels. These constraints imposed on pixel-wise segmentation justified the choice for our pipeline. It assigns additional weight to spatially continuous pixels (each weed and crop plant is a closed-loop structure) and allows us to force the minimum number of clusters to two (background and vegetation). The pixel-wise segmentation is iterated until one of the following two conditions is met : (1) the majority of pixels are classified into two clusters or (2) maximum iterations are reached. This further avoids under or over-segmentation and places an upper bound on time taken to converge to the final segmentation result. Once the segmented image is generated, the cluster with the lower number of pixels is considered to be the vegetation mask. This is based on the assumption that the number of background pixels will be greater than the vegetation pixels. Validity of this assumption is discussed in the Section \ref{section:dataset}.

In order to maximize the performance of the unsupervised segmentation, we tune the parameters of the network, purposing a randomly sampled subset ($\sim$ 30\%) of both the datasets for validation. The parameters tuned for the network are learning rate, number, and compactness of superpixels. Only one parameter is varied at a time to determine the optimal values. Other parameters are kept constant during this time. The optimal values are selected for which a maximum mean intersection-over-union (mIOU) is obtained. The experimentally determined values of the parameters are : (1) learning rate - 0.1, (2) number of superpixels - 2500, (3) compactness of superpixels - 25. Using the optimal values, the vegetation masks, $I_{veg}$ are generated for the images in the test split. As a benchmark and to compare the performance of the unsupervised approach with a supervised approach, we also trained U-Net \cite{27} on the training split of the datasets. U-Net has been shown to be an effective supervised learning approach for pixel-wise segmentation in different use cases such as medical image segmentation and autonomous driving. The network uses an encoder-decoder structure first to contract (downsample) the image and then expand (upsample) to get the final prediction. At each ``upsampling'' step, the feature map from the corresponding contraction step is also concatenated. This concatenation helps the network to learn from the lost features during downsampling. The network was utilized to predict binary class labels, with the foreground pixels (white) representing the vegetation coverage.

\subsection{Tile Classification}
\label{method_feature}

\begin{figure*}[t!]
    \includegraphics[width=\textwidth, height=0.5\textwidth]{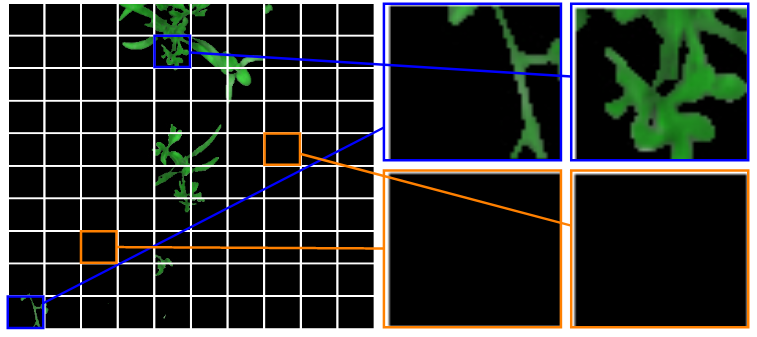}
    \caption{$I_{masked}$ is divided into smaller tiles ($I_{tile}$) as shown. Note that $I_{tile}$ is enlarged for better visualization. Also, the outline colors represents the classification for the region: Blue - Vegetation, Orange - Background/Soil.}
    \label{fig:tiles}
\end{figure*}

Once the vegetation mask $I_{veg}$ is generated, the input image $I_{RGB}$ is overlaid with $I_{veg}$ resulting in the masked image $I_{masked}$. This masked image contains only the RGB pixels for the vegetation (crops and weeds), which ensures that classification is performed based on vegetation features alone. Further, the masked image $I_{masked}$ is divided into multiple non-overlapping sub-images/tiles $I_{tile}$ of size 50x50 sq. pixels. It is possible that many regions contain a very small number of vegetation pixels or even none. Therefore, $I_{tiles}$ where vegetation coverage (number of vegetation pixels) is less than 10\% of the total area of the region (in pixels) are considered to be not infested with weed plants and are ignored in the following steps. Figure \ref{fig:tiles} shows the image $I_{masked}$ as well as the regions selected for training the classifier and the ones discarded because no vegetation pixels are present.

A variety of machine learning algorithms such as Support Vector Machines, Random Forest Classifiers, Gaussian naive Bayes, and multilayer perceptron networks \cite{svm, random_forest, nn_bayes} have commonly been used for classification. We compare the performance of these classifiers in terms of classifying $I_{tile}$ as either weed or crop. This section first discusses classifiers which uses feature vectors for classifying $I_{tile}$  as weed or crop. Also, an image based classifier is discussed which does not explicitly computes the feature vector but rather uses a fine-tuned CNN to classify $I_{tile}$  as weed or crop. 

\paragraph{Feature vector based classification:} The feature vectors are computed from the filtered set of $I_{tile}$. However, instead of extracting features based on the biological morphology, physical appearance of the crop/weed, a pre-trained CNN is utilized. The use of pre-trained CNN as a feature extractor eliminates the need for designing hand-crafted features and can be extended to any crop/weed combination. ResNet50 \cite{ResNet50} is utilized in our method to extract feature from $I_{tile}$. ResNet50 is a CNN based supervised learning approach for image classification. ResNet50 consists of multiple residual blocks stacked one after another. These residual blocks use skip connection wherein the activation of a $(l+2)^{th}$ layer is computed by the addition of activation of $(l+1)^{th}$ and $l^{th}$ layers. This skip connection helps in designing a deeper network by alleviating the problem of vanishing gradient. 

In this work, pre-trained weights learned on the ImageNet dataset \cite{23} are used. Feature vectors, of length 16,384, are extracted from one of the intermediate layers in the network (the 3rd block from the end). This is based on the intuition that the network would learn generic features such as corners and edges in the earlier layers to extract useful information while using class-specific information like shape and color in the deeper layers to label images. Principal Component Analysis (PCA) \cite{pca} is further used to reduce the vector dimensionality to 2048. PCA uses an orthogonal transformation on the set of all feature vectors to convert any possibly correlated features to a set of linearly uncorrelated features. After the feature vector from all the regions $I_{tile}$ have been extracted and reduced, they are then passed to the classification block. 

\begin{figure*}[!t]
   \centering
   \includegraphics[width=0.8\textwidth, height=0.4\textwidth]{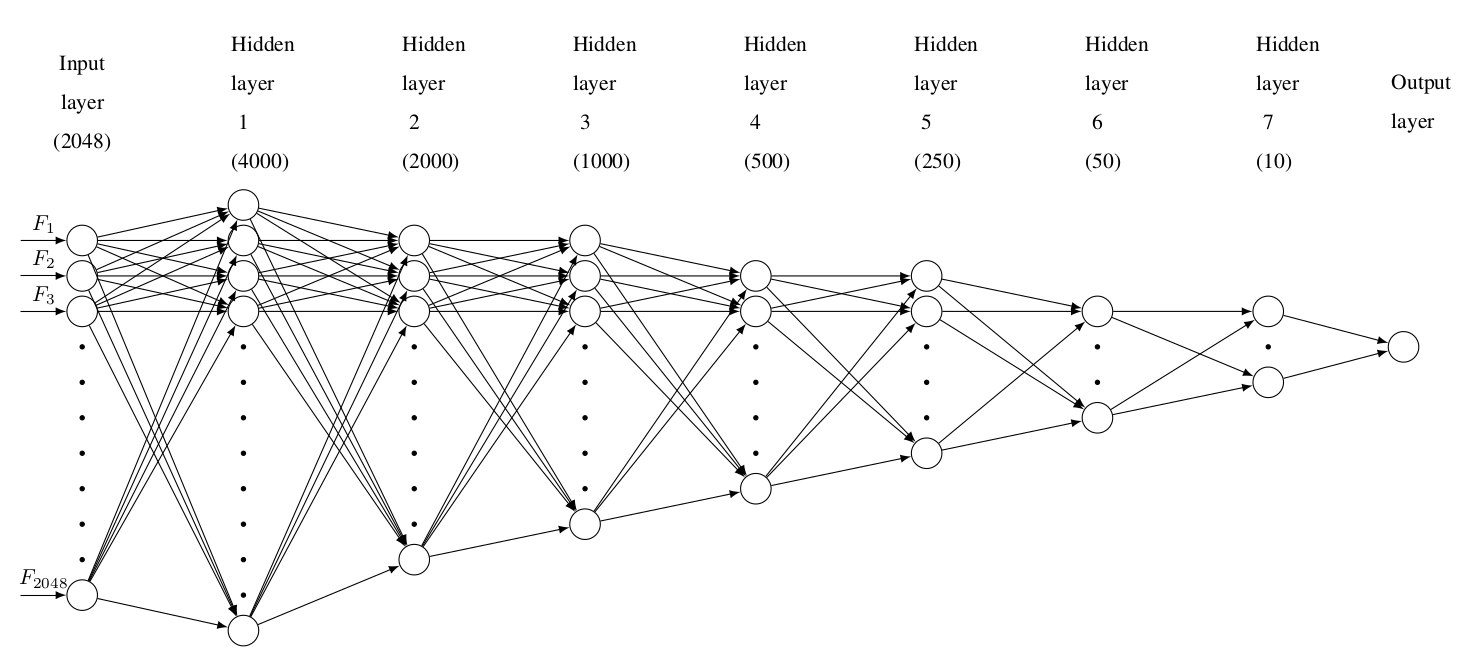}
    \caption{Multi-layer perceptron for  classifying the sub-image $I_{tile}$ as crop or weed. The input 2048 dimensional feature vectors are shown as $F_{i}$. Number of neurons for each layer are included in parentheses.}
    \label{fig:mlp}
\end{figure*}

The SVM classifier was trained with different kernels: radial basis function, linear kernel, sigmoid, and various degrees of polynomial curves. A small multi-layer perceptron was also trained for classification (Figure \ref{fig:mlp}) . It consisted of six hidden layers besides the input and output layer utilizing the Rectified Linear Unit (ReLU) activation (\cite{relu}). The activation $\textbf{a}^{[i]}$ for a neuron in $i^{th}$ layer is defined as 

\begin{eqnarray}
    \textbf{a}^{[i]} = g(\textbf{w}^{[i]}\textbf{a}^{[j]} + b^{[i]}) \\
    g(x) = max(0,x) 
    \label{formula:relu}
\end{eqnarray}

where $\textbf{w}^{[i]}$ are the weights, $\textbf{a}^{[j]}$ are the activation of the previous ($j^{th}$) layer, and $g(x)$ is the non-linear ReLU activation. The input size of the network was 2048 (length of reduced feature vectors), and the output layer consisted of a single neuron that made binary predictions (0=crop, 1=weed).

Disproportionate occurrence of crop and weed plants in the dataset leads to class imbalance in the training set. It could severely hinder the classifier's ability to recognise the weed-infested regions. In order to address the issue, we implement and compare two different sampling techniques to increase the frequency of weed plant samples as for most imbalanced data sets, the application of sampling techniques does indeed aid in improved classifier accuracy\cite{feng2019fetal}. Firstly, a combination of random oversampling  (resample certain data points from the minority class) and undersampling (drops data points from the majority class) is used while training the model, thus increasing the ratio of weed tile samples in the dataset. The alternative approach is to implement Synthetic Minority Oversampling Technique (SMOTE) \cite{28} which uses K-Nearest Neighbours algorithm to generate synthetic samples of the minority class by utilizing the existing minority class data points. Both these techniques expose the classifier to a greater number of weed plant tiles. This reduces the bias towards the majority class during the learning process.

\paragraph{Image-based classification} Instead of extracting feature vectors from the tiles, ResNet50 itself can be fine-tuned on the filtered set of tiles ($I_{tile}$) to make label predictions (crop/weed). It has been shown that ResNet50 trained on more than a million training images from ImageNet database for more than 1000 categories learns rich feature representations \cite{feature_transfer}. Same pre-trained weights described in the previous paragraph are used here as well. However, instead of using ResNet50 as a fixed feature extractor, the weights of the last layer of ResNet50 is fine-tuned. In this approach, weight of only the last layer of the pre-trained ResNet50 is updated via backpropogation, while the weights of all the other layers are fixed (freezed). In order to address the class imbalance problem, ResNet50 is tuned using the weighted binary cross-entropy loss function. The loss function for each class is defined as follows:

\begin{equation}
\label{loss}
loss[x,c] = \sum_{n} -weight[c]*(x[c] + \log{(\sum_{j}\exp{(x[j])})} ) 
\end{equation}

Here, $c$ denotes the class, $j$ $\in$ $[1,\text{number of class}]$, $n$ is the number of images in the batch and $x$ is the distance between the target and predicted label. The weights for crop and weed class are determined experimentally as 0.33 and 0.67 respectively. Besides, the performance of weighted cross-entropy loss is also compared with standard cross-entropy loss (i.e. with weights equal to 0.5). The model is trained for 250 epochs with the learning rate set to 0.001. It is also important to note that network is trained using the masked tile images instead of complete RGB images - this allows the network to make predictions based on vegetation coverage in the region instead of background/foreign objects (which are segmented out in the previous step).

\subsection{Weed density estimation}
\label{method:weed_density}

Once the weed-infested regions ($I_{tile}$ classified as a weed) have been identified, the weed density can be computed from the vegetation coverage in each individual region. In this paper, the cluster rate (\cite{wd-1}), denoted by CR, is used to quantify or model the weed density. It is defined as follows:

\begin{equation}
    \label{eqn:cluster_rate}
    CR = \frac{\mbox{Weed plant coverage in the region (in pixels)}}{\mbox{Total land area of the region (in pixels)}}
\end{equation}

The weed density estimate is crucial information in the site-specific weed management system \cite{wang_overview}. This density estimate would assist in selecting the appropriate regions for weeding with herbicides in the field. This decision-making process would depend on a variety of factors, including but not limited to crop and weed plant species and plant spacing. 

%Weed density estimation for a given region can be used to make a decision regarding where the agrochemicals should be sprayed. It might be necessary only to treat the areas where the cluster rate is above a certain threshold. This threshold would depend on a variety of factors including but not limited to crop and weed plant species as well as plant spacing. However, the selection of this threshold value is beyond the scope of this study. Here, we are only concerned with the robust estimation of the cluster rate corresponding to the weed-infested regions.

\subsection{Dataset}
\label{section:dataset}

The proposed approach is validated on two commonly used publicly available datasets: Crop/Weed Field Image dataset \cite{cwfid_dataset} and the Sugar Beets dataset \cite{sugar_dataset}. Authors of both the datasets provide annotated images that mark crop and weed plant pixels distinctly.

\paragraph{Crop/Weed Field Image dataset (CWFID)} It contains images acquired by an autonomous field robot BoniRob from a carrot farm. This dataset includes 60 top-down field images with intra-row and close-to-crop weeds. In this work, CWFID was augmented using common techniques such as skewing, flipping, rotating, and zooming. This resulted in a total of 90 images, which are split into train and test set in the ratio of 2:1 (60 training images, 30 testing images). 

\paragraph{Sugar Beets dataset} It contains field images acquired by the same autonomous robot BoniRob from a sugar beet farm for over three months. While the entire dataset is quite extensive and includes data from multiple sensors, only a subset of the Sugar Beets dataset (500 images) is used in this study. Compared to CWFID, the dataset presents a variation in terms of plant species and the number of overlapping plants. Besides, unlike CWFID, Sugar Beets dataset suffers from poor contrast arising due to insufficient illumination. It is further split into the train and test set with a ratio of 7:3 (350 training images, 150 testing images). 

\paragraph{Classification dataset} 
Since the pixel wise annotated masks were provided with the dataset, the tile label for classification was deduced from this pixel level annotation. The full sized images were divided into 50x50 squares and a corresponding label was chosen for the tile based on the number of the crop/weed pixels in the tile image. If the tile image contained more crop pixels, it was labelled as crop, if it contained more weed pixels it was labelled as weed. Any tiles with less than 10\% vegetation coverage were ignored. Table 1 shows the number of crop and weed tiles used by the authors to train the model from the dataset. It is important to note that in absence of pixel wise annotations, only a single binary label for each tile needs to be manually specified. This will reduce the time and effort required for the annotation process significantly.
\begin{table}[!ht]
    \centering
    \begin{tabular}{|c|c|c|c|c|}
        \hline
        Dataset & \multicolumn{2}{c|}{Crop} & \multicolumn{2}{c|}{Weed} \\
        \hline
        & Train & Test & Train & Test \\
        \hline
        CWFID & 1370 & 411 & 637 & 244 \\
        \hline
        Sugar Beets & 5585 & 2833 & 2555 & 947\\
        \hline
    \end{tabular}
    \caption{Tiles generated for Classification.}
    \label{tab:tile_dataset}
\end{table}

Since both the datasets are captured from ground robots, the vegetation pixel density in the images is sparse compared to background or soil pixel density. The vegetation pixel density for CWFID and Sugar Beets dataset is 10.09\% and 6.58\%, respectively. Further, the maximum vegetation density in a single image for the dataset is 23.76\% for CWFID and 19.53\% for the Sugar Beets dataset. Hence, due to the steep difference in the vegetation pixel and soil pixel densities, the assumption made in Section \ref{method_segmentation} is reasonable (the vegetation cluster will always contain a lower number of pixels compared to the background).

The motivation behind selecting datasets captured using the same autonomous robot was to facilitate easier integration of the proposed method in the existing infrastructure. Even though the images have been acquired using the same autonomous robot, the two dataset contains two different crop/weed species (Sugar Beets and Carrots Crops) with varying overlap and image contrast (likely due to varying lighting/illumination). This results in a significant difference in the image content of the two datasets.

\subsection{Evaluation}
\label{method_eval}

Mean intersection-over-union (mIoU) is a popular metric to evaluate pixel-wise segmentation networks \cite{17}. However, the focus of our approach is not a dense segmentation prediction. Instead, the accuracy of weed distribution and density estimation is evaluated as explained below. Besides, in order to justify the choice of individual components, we also evaluate the output from the intermediate step (unsupervised binary segmentation).

\paragraph{Vegetation segmentation} The first step of the pipeline segments the vegetation pixels for the input RGB image. The vegetation pixels are denoted by white color in the binary image. 
Ground truth vegetation coverage ($coverage_{GT}$) and predicted vegetation coverage ($coverage_{pred}$) are compared using the mIoU metric defined below. It helps determine the effectiveness with which the vegetation pixels are identified in the image. 

\begin{equation}
\label{iou}
mIoU = \frac{\sum_{i}x_{ii}}{C\left(\sum_{i}\sum_{j}x_{ij}+\sum_{j}x_{ji}-x_{ii}\right)}
\end{equation}

where $C$  is the number of classes (two in this work), $x_{ij}$ represents the number of pixels belonging to class $i$ and predicted as class $j$. The maximum value of mIoU is 1.0 which signifies that the all the pixels are correctly labelled.

\paragraph{Weed distribution estimation} The predictions made by the classification model for all the regions (squares tiles) are compared against the ground truth labels. Classifier accuracy highlights the overall system performance but is also biased towards the majority class (the class with a significantly greater number of samples). Hence, the improvements in the classifier performance with respect to a specific class can be reflected by the recall and precision metrics (Equations  \ref{eqn:precision}, \ref{eqn:recall}). Besides, the F1-score can be utilized to study the overall performance of the classifier (Equation \ref{eqn:F1Score})

% For each classifier, the performance is quantified by computing accuracy using Equation \ref{eqn:accuracy}.

% \begin{equation}
%     \label{eqn:accuracy}
%     Accuracy = \frac{\mbox{Number of tiles classified correctly}}{\mbox{Total number of tiles}} 
% \end{equation}

\begin{equation}
    \label{eqn:precision}
    precision=\frac{TP}{TP+FP} 
\end{equation}

\begin{equation}
    \label{eqn:recall}
    recall=\frac{TP}{TP+FN} 
\end{equation}

\begin{equation}
\label{eqn:F1Score}
F1-score = \frac{2 \cdot {precision}\cdot {recall}}{{precision}+{recall}}
\end{equation}

Here, TP refers to true positives, FP refers to false positives, and FN refers to false negatives. For a particular class, true positives are all the regions correctly assigned to that class, false positives are the regions incorrectly assigned to that class, and false negatives are the regions incorrectly labeled as another class. While F1-score is an essential and reliable metric for measuring the classifier performance, recall is also afforded significant weightage. The reason for our choice is that we aim to minimize misclassification for weed-infested regions. 
% However, it should be noted that care was taken that recall weightage did not render the other two metrics insignificant.

\paragraph{Weed density} In order to evaluate weed density, error in predicted density for each region correctly classified is computed. The weed density for each tile is measured by the cluster rate (Equation \ref{eqn:cluster_rate}). The estimated cluster rate ($CR_{est}$) is compared against cluster rate in ground truth pixel-wise annotations ($CR_{gt}$). The following three metrics are computed to quantify the error in weed density estimation: (1) mean accuracy, (2) mean absolute error (MAE) and (3) root mean squared error (RMSE) (Equations \ref{eq:mean_accuracy}, \ref{eq:mae}. \ref{eq:rmse}). 

\begin{equation}
    \label{eq:wd_error}
    \centering
    \mbox{Absolute Error} = \mid \mbox{$CR_{gt}$} -  \mbox{$CR_{est}$} \mid
\end{equation}

\begin{equation}
    \centering
    \mbox{Mean Accuracy} = 1 - \sum_{i}{
    \frac{\mbox{Absolute Error} / \mbox{$CR_{gt}$}}{N}
    }
    \label{eq:mean_accuracy}
\end{equation}
    
\begin{equation}
    \centering
    \mbox{MAE} = \sum_{i}{\frac{\mbox{Absolute Error}}{\mbox{N}}}
    \label{eq:mae}
\end{equation}
    
\begin{equation}
    \centering
    RMSE = \sqrt{ \frac{
    \sum_{i} \mbox{Absolute Error} ^{2}}{N}
    }     
    \label{eq:rmse}
\end{equation}
    
where $CR_{i}$ is the ratio of weed plant coverage in the given tile to the total land area of the region (both in pixels), $i$ = {Ground Truth (gt), Estimated (est)} and $N$ is the total number of regions/tiles. 

\begin{figure*}
\centering

    \begin{subfigure}
    
        \includegraphics[width=0.2\textwidth, height=0.18\textwidth]{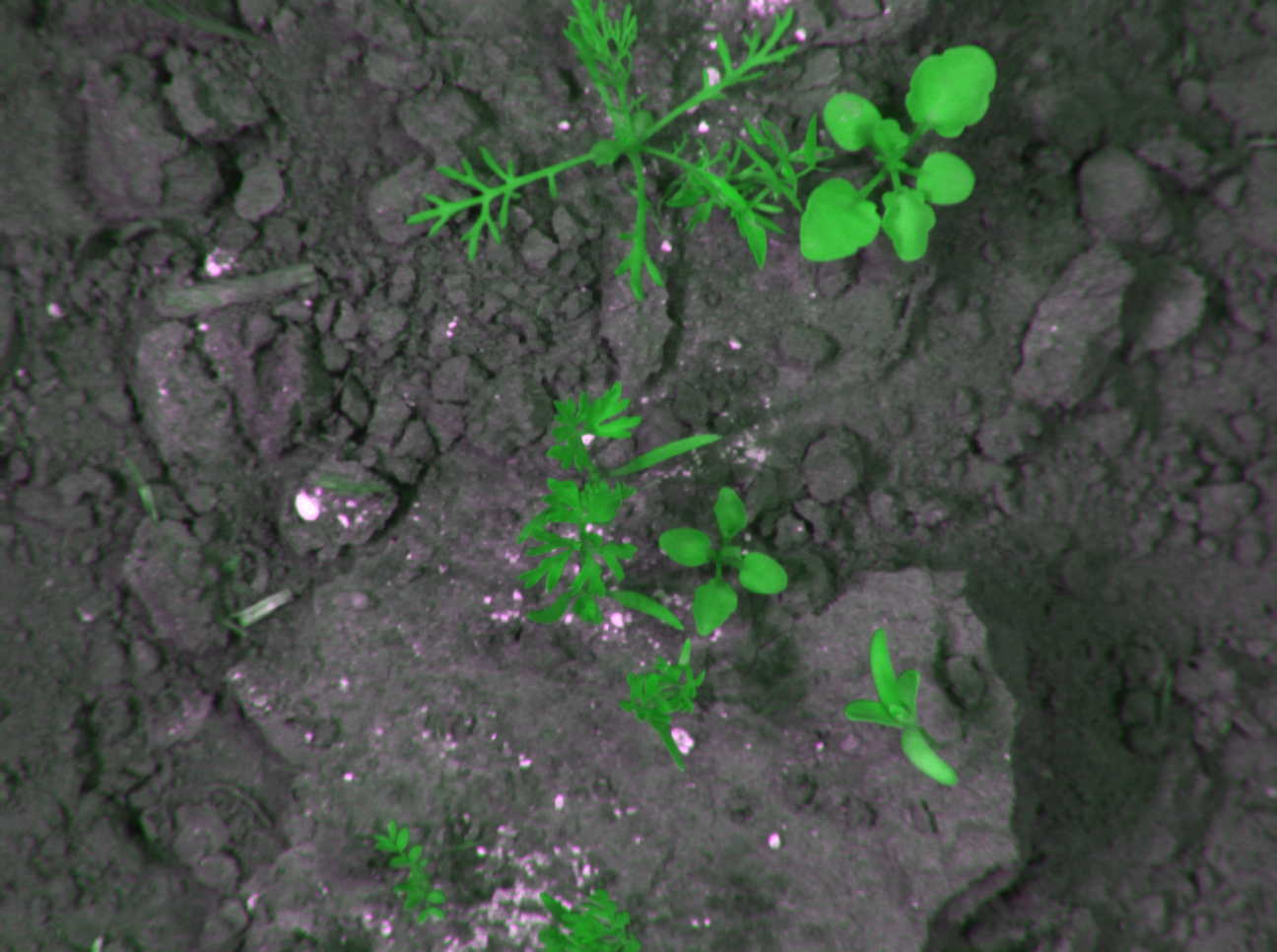}
        \hspace{0.25cm}
        \includegraphics[width=0.2\textwidth, height=0.18\textwidth]{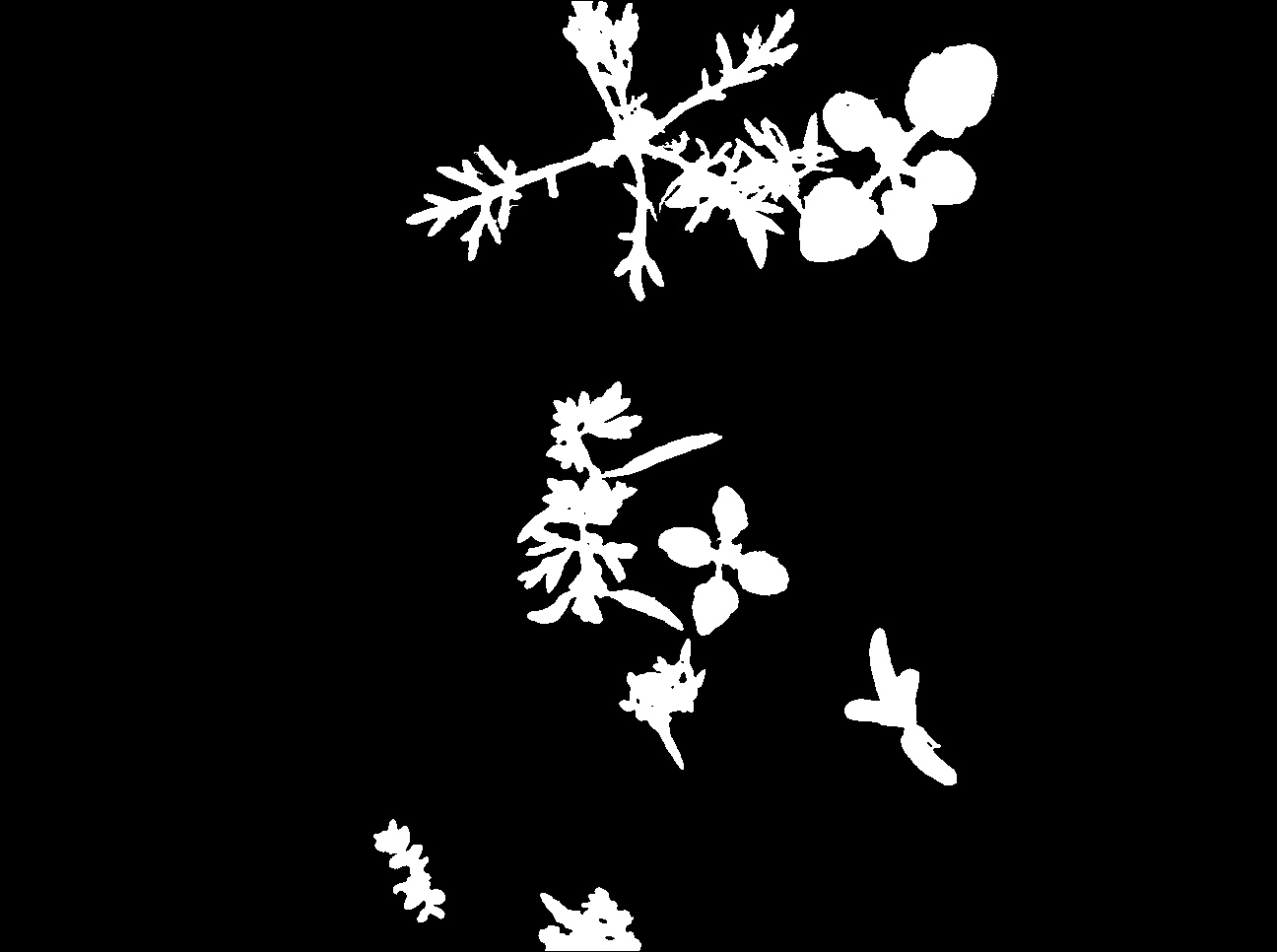}
        \hspace{0.25cm}
        \includegraphics[width=.2\textwidth, height=0.18\textwidth]{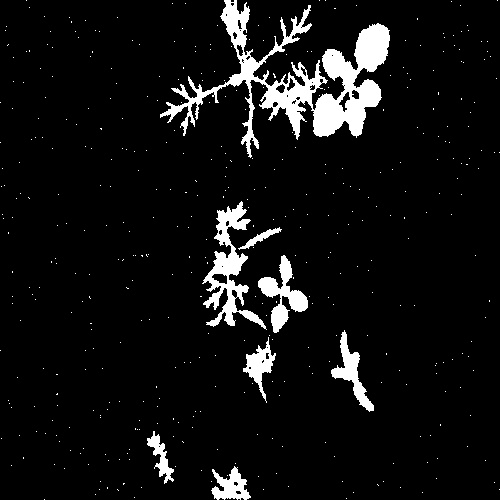}
        \hspace{0.25cm}
        \includegraphics[width=.2\textwidth, height=0.18\textwidth]{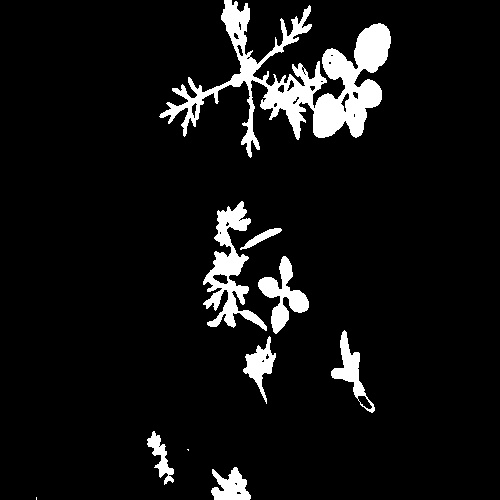}
    
        \vspace{0.35cm}
        
        \includegraphics[width=0.2\textwidth, height=0.18\textwidth]{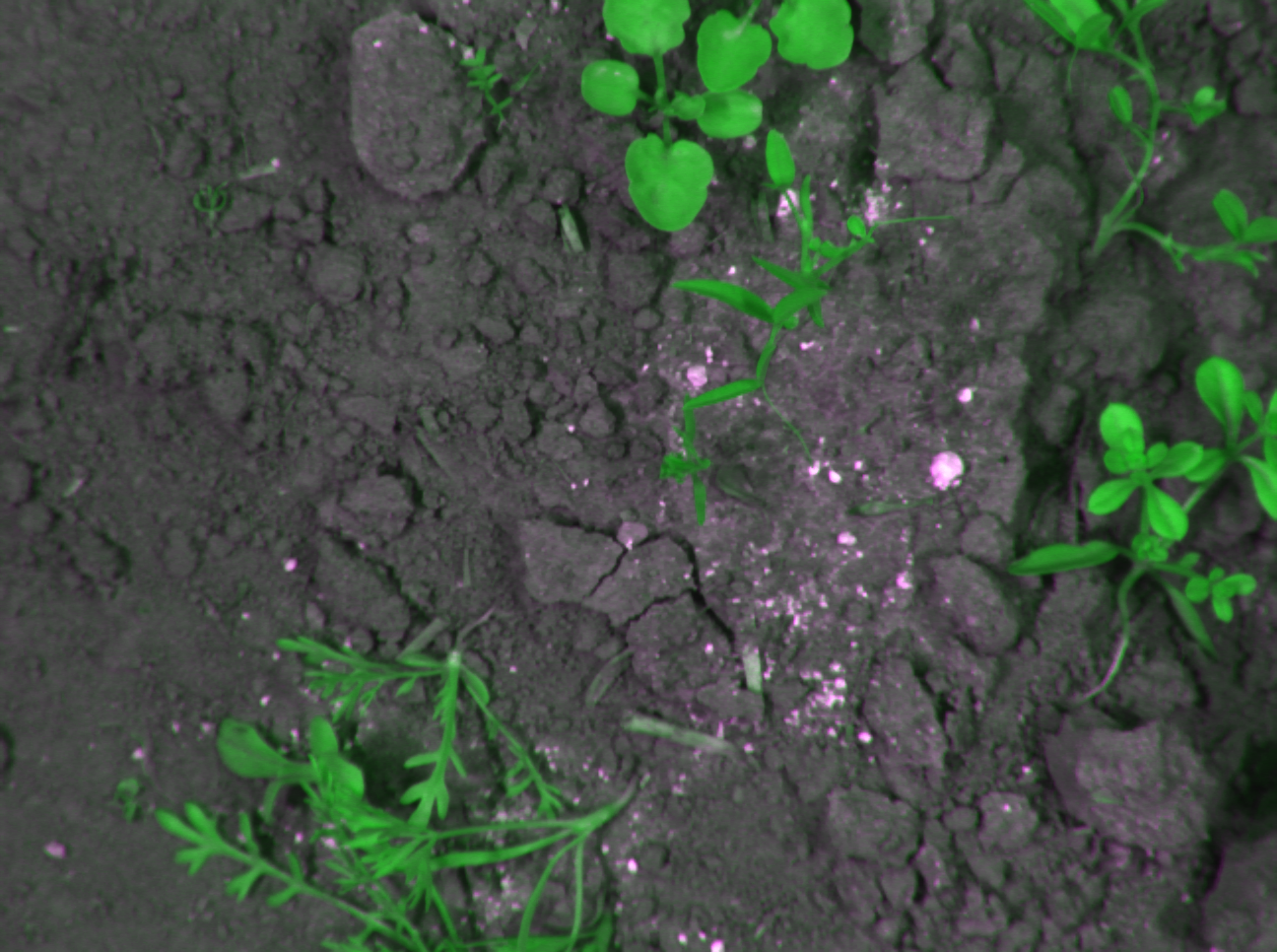}
        \hspace{0.25cm}
        \includegraphics[width=.2\textwidth, height=0.18\textwidth]{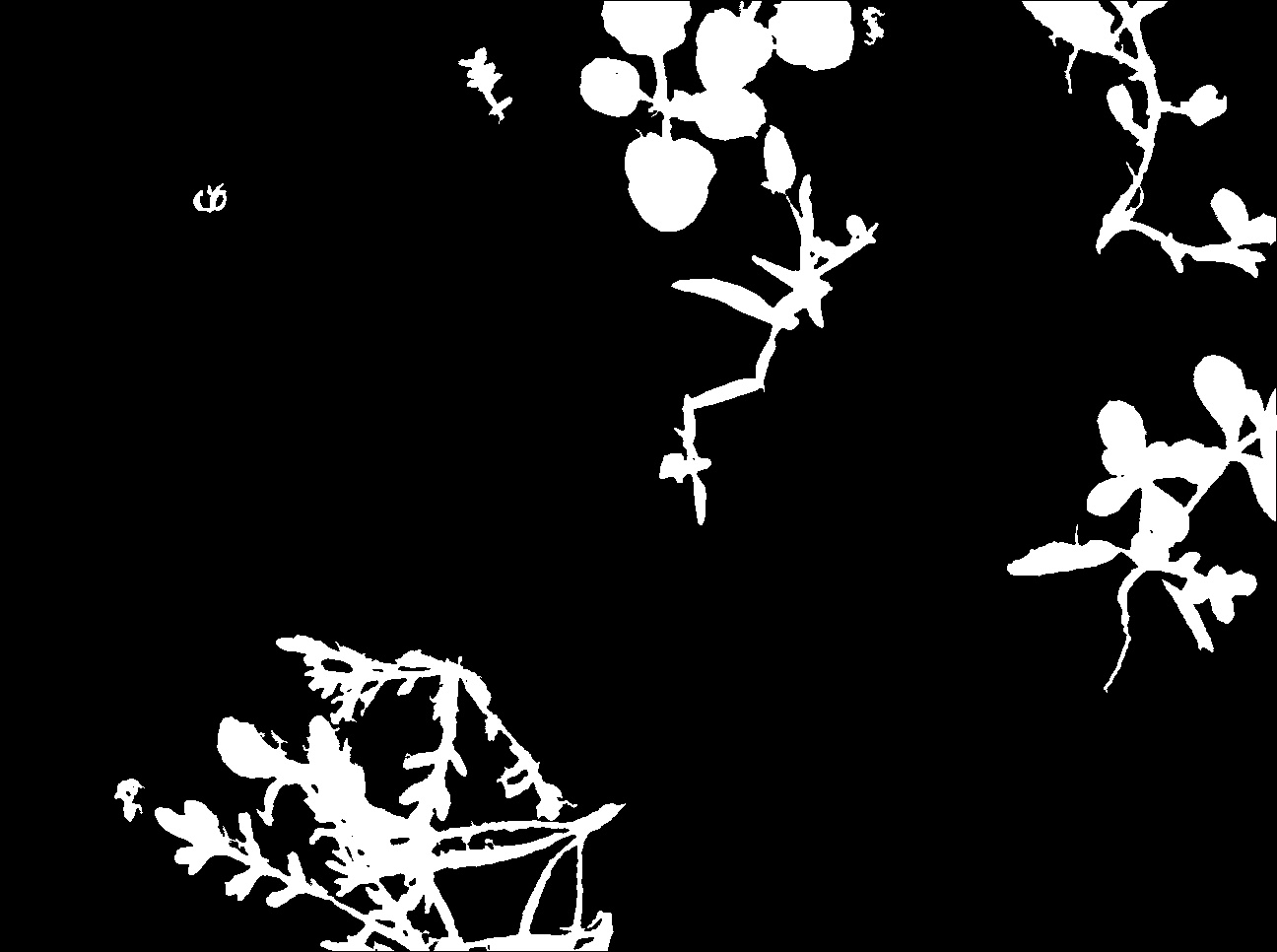}
        \hspace{0.25cm}
        \includegraphics[width=.2\textwidth, height=0.18\textwidth]{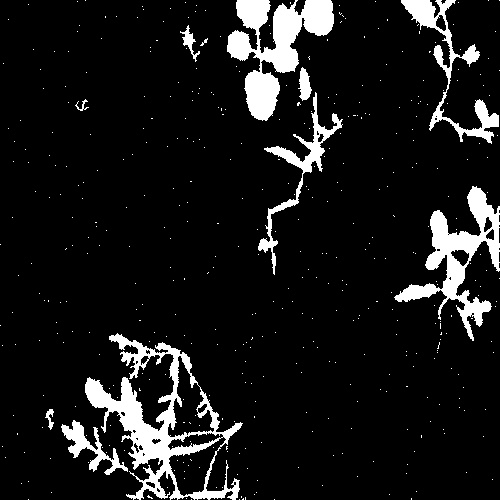}
        \hspace{0.25cm}
        \includegraphics[width=.2\textwidth, height=0.18\textwidth]{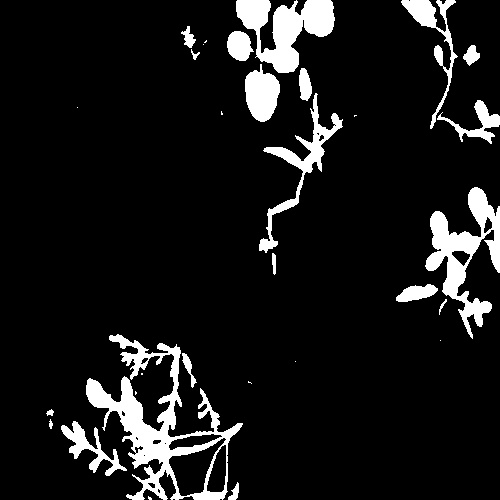}
        
        \vspace{0.25cm}
        (a) CWFID
        
    \end{subfigure}
    
    \vspace{0.5cm}
    
    \begin{subfigure}
            
        \includegraphics[width=0.2\textwidth, height=0.18\textwidth]{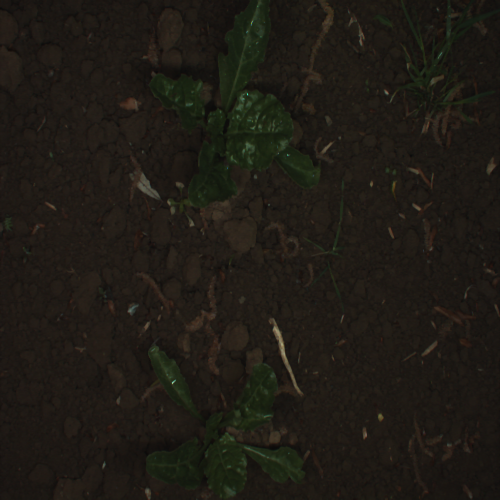}
        \hspace{0.25cm}
        \includegraphics[width=.2\textwidth, height=0.18\textwidth]{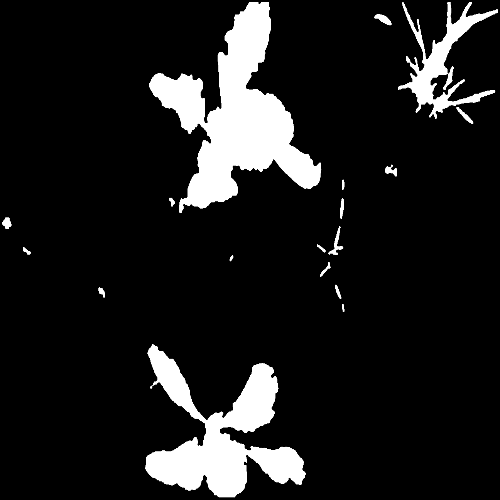}
        \hspace{0.25cm}
        \includegraphics[width=.2\textwidth, height=0.18\textwidth]{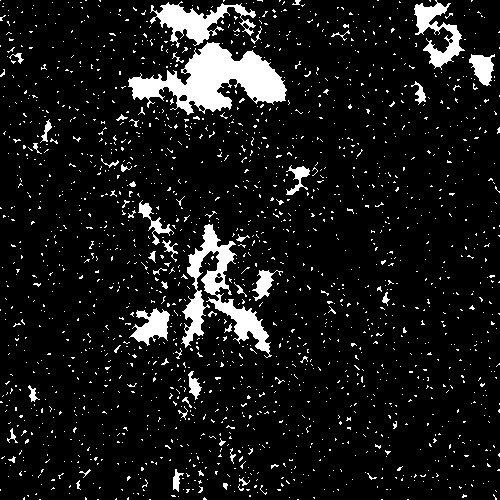}
        \hspace{0.25cm}
        \includegraphics[width=.2\textwidth, height=0.18\textwidth]{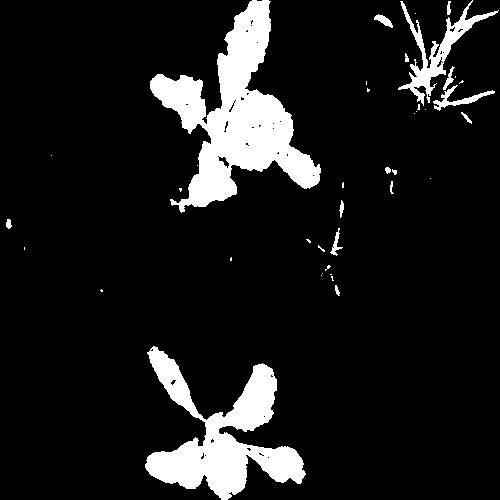}
    
        \vspace{0.35cm}
        
        \includegraphics[width=0.2\textwidth, height=0.18\textwidth]{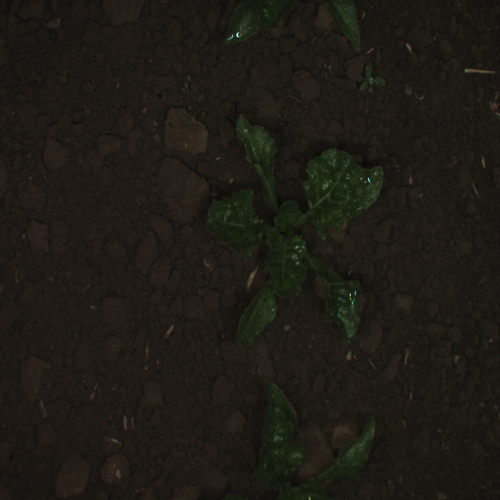}
        \hspace{0.25cm}
        \includegraphics[width=.2\textwidth, height=0.18\textwidth]{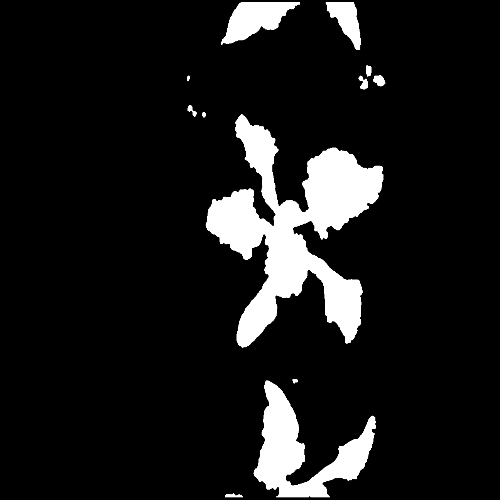}
        \hspace{0.25cm}
        \includegraphics[width=.2\textwidth, height=0.18\textwidth]{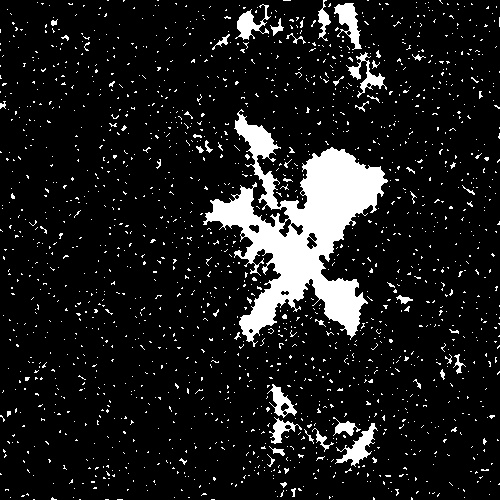}
        \hspace{0.25cm}
        \includegraphics[width=.2\textwidth, height=0.18\textwidth]{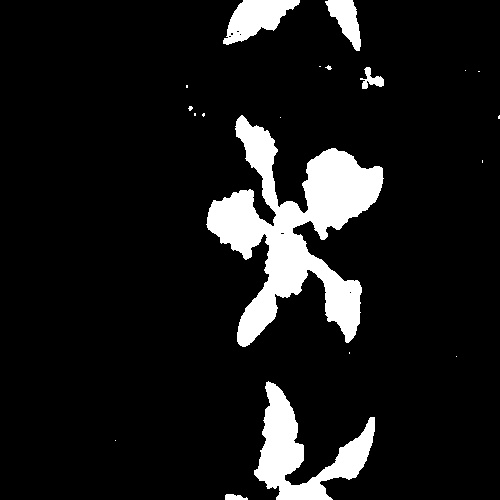}
        
        \vspace{0.25cm}
        (b) Sugar Beet Dataset
        
    \end{subfigure}
        
    \caption{Vegetation masks (left to right) : (I) Original Image, (II) Ground truth vegetation mask, (III) Vegetation mask predicted by UNet, (IV) Vegetation mask predicted by the unsupervised network.}
    \label{fig:veg_masks}
\end{figure*}

\subsection{Software}
\label{software}

The proposed approach is developed using the Python programming language. Most of the programs are developed from scratch by the authors while open-source implementations are also used. In addition, common libraries such as OpenCV \cite{opencv} and Scikit-Learn \cite{sklearn} are also utilized. The program is developed using Ubuntu 16.04 LTS operating system. The system had an octa-core CPU (Intel  i7-7700HQ) and a NVIDIA GeForce GTX 1050 Ti (4GB RAM) graphics card. The code will be made available publicly on GitHub at \url{https://github.com/ShantamShorewala/weed-distribution-and-density-estimation}. 

%%%%%%%%%%%%%%%%%%%%%%%%%%%%%%%%%%%%%
\section{Results and discussion}

\subsection{Vegetation segmentation}
\label{SubSec:ResultsVegetationSegmentation}
The performance of the CNN based unsupervised segmentation for vegetation segmentation using the method presented in Section \ref{method_segmentation} is described in this section.

\paragraph{Qualitative evaluation} Figure \ref{fig:veg_masks} visualizes the vegetation segmentation results for a few input images from both the datasets. It can be inferred that the unsupervised segmentation network matches, and in some cases even outperforms supervised segmentation (U-Net) in terms of segmenting the vegetation pixels from the background.  For instance, the unsupervised segmentation approach is able to segment the ﬁner details of the vegetation, as shown in Figure \ref{fig:veg_masks} (Top right corner of the image in the third row). It is also interesting to note that U-Net also labels some individual or extremely small clusters of pixels as vegetation. The unsupervised approach, in general, does not repeat this trend. It can be attributed to the fact that it gives weightage to spatial continuity of the vegetation clusters, whereas U-Net tends to look at only the immediate surroundings of a pixel (downsampling using max pooling). This trend was much more prominent for the Sugar Beets dataset, which exhibits poor contrast in comparison with the CWFID.

\paragraph{Quantitative evaluation} Mean intersection-over-union (mIoU) computed for both the test split of both the datasets is used to compare the performance of the two networks. The results are reported in Table \ref{tab:seg_comparison}. It is observed that the unsupervised network produced a slightly higher score for the CWFID and significantly outperformed U-Net for the Sugar Beets dataset. This observation can be attributed to the fact that, unlike U-Net, the unsupervised approach does not need to learn mapping from specific features to discriminate between the vegetation and background pixels. It should also be noted that both supervised and unsupervised approaches perform better for the CWFID compared to the Sugar Beets dataset. This observation can be attributed to the fact that the Sugar Beets dataset's images suffer from poor illumination and, hence, poorer contrast. These results justify the selection of the unsupervised network to segment the vegetation pixels from images captured for different plant species under varying conditions.

\begin{table}[!ht]
    \centering
    \caption{Quantitative evaluation for vegetation segmentation.}
    \begin{tabular}{|c|c|c|}
        \hline
        Model & Dataset & mIoU\\
        \hline
        Unsupervised Segmentation & CWFID & 0.928 \\
        \hline
        Unsupervised Segmentation & Sugar Beet Dataset & 0.82 \\
        \hline
        UNet & CWFID & 0.913 \\
        \hline
        UNet & Sugar Beet Dataset & 0.76 \\
        \hline
    \end{tabular}
    \label{tab:seg_comparison}
\end{table}

\subsection{Feature vector based tile classification}
As discussed earlier, the vegetation segmentation $I_{veg}$ is utilized to identify the regions of vegetation containing crops and weed in the images. Subsequently, the input image $I_{RGB}$ is overlaid with $I_{veg}$ resulting in masked image $I_{masked}$. This masked image is then divided into non-overlapping tiles (sub-images) $I_{tile}$. Subsequently, the features are extracted from each $I_{tile}$ using a pre-trained ResNet50. The performance of different classifiers in classifying $I_{tile}$ as weed or crop using these features is presented in Table \ref{tab:classifier_comparison}. 

Note the improvement in classifier performance due to weighted training using different approaches. (Table \ref{tab:classifier_comparison}). This result substantiate previous finding \cite{28} by demonstrating that sampling techniques (random sampling and SMOTE) helps in improving the classifier performance for an unbalanced dataset. The performance is measured using the computed precision and recall for the weed class on the test set. While the precision and recall values improve relatively due to sampling techniques that address class imbalance, the absolute values remain below the acceptable threshold. Random forest classifier achieved a recall of 1.0 but an extremely poor precision - all tiles were predicted as weed-infested. This demonstrates the inability of these classifiers to robustly discriminate between feature vectors generated from the proposed pipeline corresponding to crop and weed plants.

\begin{table*}
    \centering
    \caption{Comparison of classifier performance on the two datasets. (LK=Linear Kernel, PK = Polynomial Kernel, GNB = Gaussian Naive Bayes, NN = Neural Network, RF = Random Forest)}
    \begin{tabular}{|c|c|c|c|c|c|c|c|c|c|c|}
        \hline
        Dataset & Classifier & \multicolumn{3}{c|}{Vanilla} & \multicolumn{3}{c|}{Random Sampling} & \multicolumn{3}{c|}{SMOTE} \\
        \hline
        &  & Precision & Recall & F1 & Precision & Recall & F1 & Precision &  Recall &F1 \\
        \hline
        \multirow{8}{*}{CWFID} & SVM with LK & 0.26 & 0.12 & 0.16 & 0.3 & 0.23 & 0.26 & 0.31 & 0.22 & 0.26 \\\cline{2-11}
        % \hline
        & SVM with 2nd order PK & 0.11 & 0.03 & 0.04 & 0.1 & 0.02 & 0.03 & 0.31 & 0.17 & 0.22\\\cline{2-11}
        % \hline
        & SVM with 3rd order PK & 0.0 & 0.0 & - & 0.33 & 0.01 & 0.02 & 0.28 & 0.15 & 0.2\\\cline{2-11}
        % \hline
        & SVM with RBF Kernel & 0.31 & 0.16 & 0.21 & 0.28 & 0.22 & 0.25 & 0.16 & 0.04 & 0.06\\\cline{2-11}
        % \hline
        & SVM with Sigmoid Kernel & 0.31 & 0.13 & 0.18 & 0.29 & 0.23 & 0.26 & 0.29 & 0.31 & 0.30\\\cline{2-11}
        % \hline
        & GNB & 0.16 & 1 & 0.28 & 0.16 & 0.1 & 0.28 & 0.16 & 1 & 0.26\\\cline{2-11}
        % \hline
        & NN & 0.32 & 0.30 & 0.31 & 0.31 & 0.31 & 0.31 & \textbf{0.25} & \textbf{0.92} & \textbf{0.39}\\\cline{2-11}
        % \hline
        & RF & 0.29 & 0.07 & 0.11 & 0.35 & 0.14 & 0.2 & 0.22 & 0.06 & 0.0.09\\
        \hline

        \multirow{8}{*}{Sugar Beets} & SVM with LK & 0.35 & 0.36 & 0.36 & 0.31 & 0.40 &0.35 & 0.26 & 0.40 & 0.32 \\\cline{2-11}
        % \hline
        & SVM with 2nd order PK & 0.32 & 0.50 & 0.39 & 0.32 & 0.37 & 0.34 & 0.31 & 0.52 & 0.39\\\cline{2-11}
        % \hline
        & SVM with 3rd order PK & 0.32 & 0.16 & 0.21 & 0.26 & 0.68 & 0.38 & 0.32 & 0.17 & 0.22\\\cline{2-11}
        % \hline
        & SVM with RBF Kernel & 0.44 & 0.13 & 0.20 & 0.28 & 0.47 &0.35 & 0.34 & 0.46 & 0.39 \\\cline{2-11}
        % \hline
        & SVM with Sigmoid Kernel & 0.34 & 0.42 & 0.38 & 0.29 & 0.59 & 0.39 & 0.29 & 0.61 & 0.39\\\cline{2-11}
        % \hline
        & GNB & 0.26 & 0.60 & 0.36 & 0.16 & 0.85 & 0.27 & 0.23 & 0.67 & 0.34 \\\cline{2-11}
        % \hline
        & NN & 0.37 & 0.21 & 0.27 & 0.33 & 0.46 & 0.38 & \textbf{0.28} & \textbf{0.70} & \textbf{0.40}\\\cline{2-11}
        % \hline
        & RF & 0.36 & 0.38 & 0.37 & 0.26 & 0.55 & 0.35 & 0.30 & 0.32 & 0.31\\
        \hline
        
    \end{tabular}
    \label{tab:classifier_comparison}
    
    % \caption{Accuracy and recall of weed class on fine tuned resnet}
    % \begin{tabular}{|c|c|c|}
    %      \hline
    %      Classifier & Accuracy & Recall\\
    %      \hline
    %      ResNet-50 & dummy & dummy\\
    %      \hline
    % \end{tabular}
    % \label{tab: classifier_comparison2}

\end{table*}

\subsection{Effect of tile size on classification accuracy}

The choice of tile size (a square with side 50 pixels) was intuitively based on two observations: (1) it resulted in regions where pixels belonged largely to either one of crop or weed plants instead of both and (2) it avoided the formation of regions with nearly all pixels belonging to vegetation cluster. This would mean there would not be enough descriptive features for the classifier to distinguish crop and weed plants since they would look extremely similar. However, to validate the choice of tile size, results from regions of different sizes were compared. In this study, we retrained the classification models by both increasing and decreasing the side length (100 and 25 pixels, respectively). Table \ref{tab:patch_size_comparison} reports precision and recall values for all the machine learning classifiers. 

\begin{table*}
    \centering
    \caption{Comparison of classifier performance for different patch sizes. (LK=Linear Kernel, PK = Polynomial Kernel, GNB = Gaussian Naive Bayes, NN = Neural Network, RF = Random Forest)}
    \begin{tabular}{|c|c|c|c|c|c|c|c|c|c|c|}
        \hline
        Dataset & Classifier & \multicolumn{3}{c|}{$50\times 50$} & \multicolumn{3}{c|}{$100\times 100$} & \multicolumn{3}{c|}{$25\times 25$} \\ \hline
        &  & Precision & Recall & F1 & Precision & Recall & F1 & Precision &  Recall & F1 \\ \hline
        \multirow{8}{*}{CWFID} & SVM with LK &0.26&	0.12& 0.16& 0.39&	0.32& 0.35&	0.35& 0.48& 0.40\\\cline{2-11}
        & SVM with 2nd order PK &0.11	&0.03 &0.04	&0.40	&0.02 &0.04	&0.0 &0.0 &- \\\cline{2-11}
        & SVM with 3rd order PL &0.0	&0.0 &-	&0.22	&0.07 &0.11	&0.0	&0.0 &- \\\cline{2-11}
        & SVM with RBF Kernel &0.31	&0.16 &0.21	&0.43	&0.07 &0.14	&0.34	&0.43 &0.38 \\\cline{2-11}
        & SVM with Sigmoid Kernel &0.31	&0.13 &0.18	&0.39	&0.22 &0.28	&0.34	&0.43 &0.38\\\cline{2-11}
        & GNB &0.16	&1.0 &0.28	&0.0	&0.0 &-	&0.35	&0.24 &0.29 \\\cline{2-11}
        & NN   &0.32	&0.30 &0.31	&0.30	&0.52 &0.38	&0.42	&0.21 &0.28 \\\cline{2-11}
        & RF &0.29 &0.07 &0.11 &0.75	&0.07  &0.13 &0.36 &0.16 &0.22 \\
        \hline 
        
        \multirow{8}{*}{Sugar Beets} & SVM with LK &0.35     &0.36 &0.36	&0.42 &0.21 &0.28 &0.37 &0.14 &0.20\\\cline{2-11}
        & SVM with 2nd order PK &0.32	&0.50  &39	&0.27	&0.42 &0.33	&0.31	&0.29 &0.30\\\cline{2-11}
        & SVM with 3rd order PL &0.32	&0.16 &0.21	&0.28	&0.09 &0.14	&0	&0 &-\\\cline{2-11}
        & SVM with RBF Kernel &0.44	&0.13 &0.20	&0.34	&0.39 &0.36	&0.29	&0.67 &0.41\\\cline{2-11}
        & SVM with Sigmoid Kernel &0.34	&0.42 &0.38	&0.31	&0.44 &0.36	&0.28	&0.7 &0.40 \\\cline{2-11}
        & GNB &0.26	&0.60 &0.36	&0	&0 &-	&0.12	&0.97 &0.21 \\\cline{2-11}
        & NN  &0.36 &0.38 &0.37 &0.35	&0.25 &0.29 &0.4 &0.13 &0.2\\\cline{2-11}
        & RF &0.37	&0.21 &0.27	&0.25	&0.22 &0.23	&0.28	&0.43 &0.34\\
        \hline
    \end{tabular}
    \label{tab:patch_size_comparison}
    
    \bigskip
    
    \centering
    \caption{Computation time for passing a single image.}
    \begin{tabular}{ |c|c|c|c| }
        \hline
        Tile side length & 100 px & 50 px & 25 px \\
        \hline
        Computation time & 0.96 s & 0.90 s & 5.22 s
        \\
        \hline
    \end{tabular}
    \label{tab:computation_time}

\end{table*}

Considering both precision and recall values, classifiers trained with tiles of side length 25 and 50, on average, outperform those trained on $100\times 100$ tiles. Further, Table \ref{tab:computation_time} reports the computation time required to pass the image through the classification block (includes time to generate the tiles, extract feature vectors, and classify them). The computation time is similar for patch sizes with side 50 and 100 pixels but increases significantly for side length 25 pixels. The reason was that the number of tiles with vegetation pixel density greater than 10\% remained similar for the first two but is much higher for patches with side of length 25 pixels. Hence, in order to choose between the tile side length between 25, 50, and 100 pixels, processing or computation time for a single image is also taken into account. Based on these considerations, choice of patch size with side of length 50 pixels is justified for both datasets. It may be noted that  the tile size needs to be computed only once and would be fixed for the entire agricultural season for a particular crop/weed species  assuming no significant changes in the acquisition system on-board the autonomous vehicle.

\subsection{Image-based classification}

\begin{figure*}
    \includegraphics[width=.18\textwidth, height=0.17\textwidth]{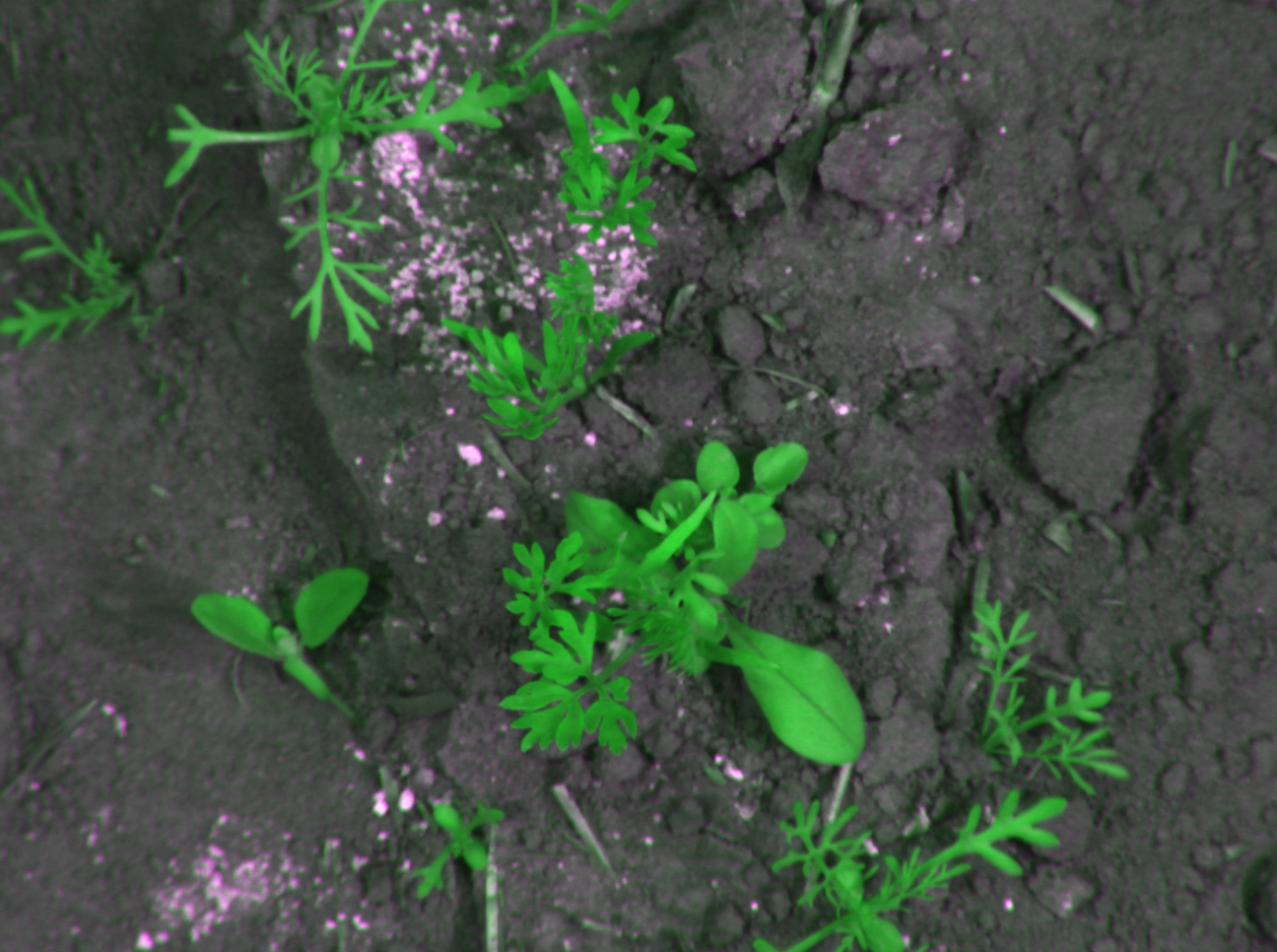}\hfill
    \includegraphics[width=.18\textwidth, height=0.17\textwidth]{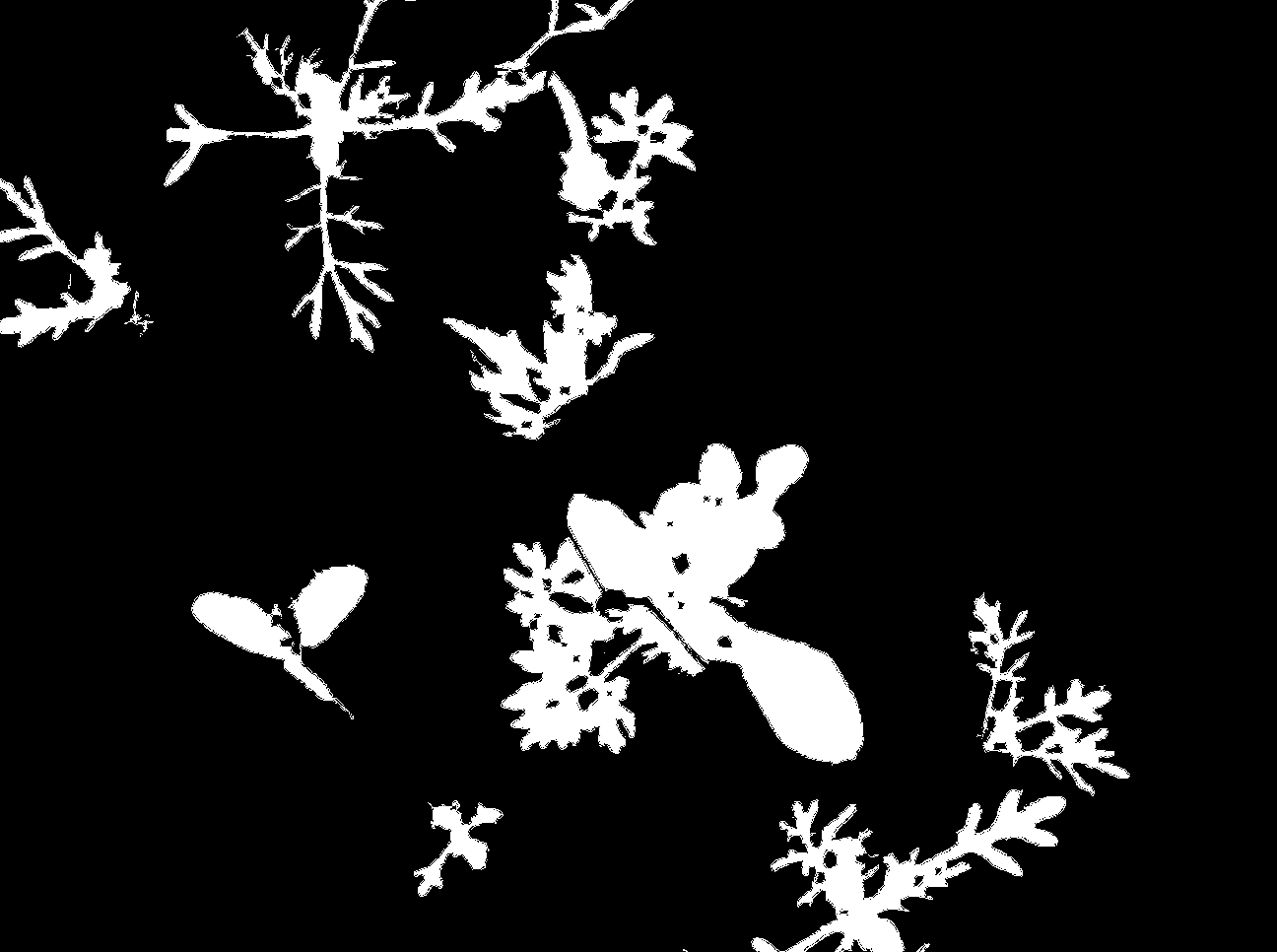}\hfill
    \includegraphics[width=.18\textwidth, height=0.17\textwidth]{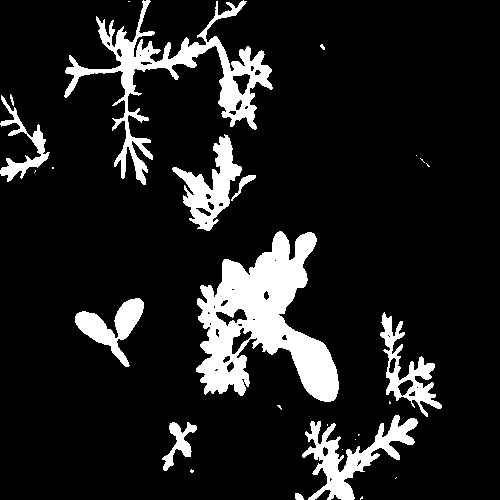}\hfill
    \includegraphics[width=.18\textwidth, height=0.17\textwidth]{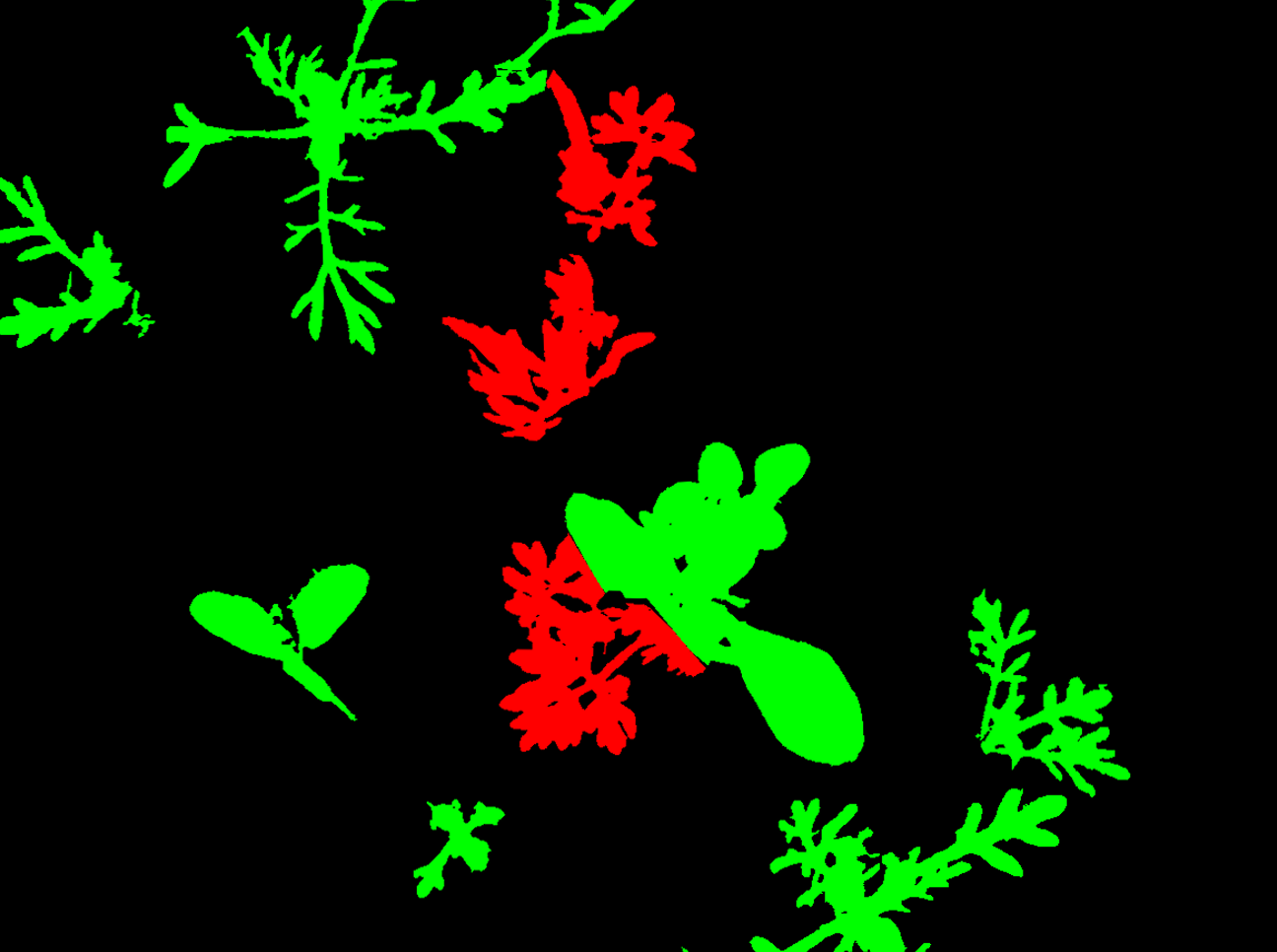}\hfill
    \includegraphics[width=.18\textwidth, height=0.17\textwidth]{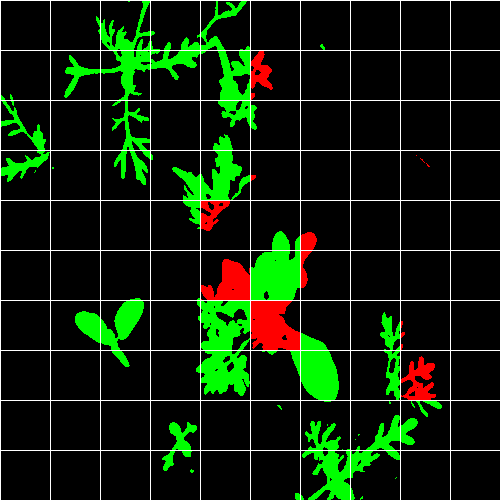}

    \vspace{0.35cm}
    \includegraphics[width=.18\textwidth, height=0.16\textwidth]{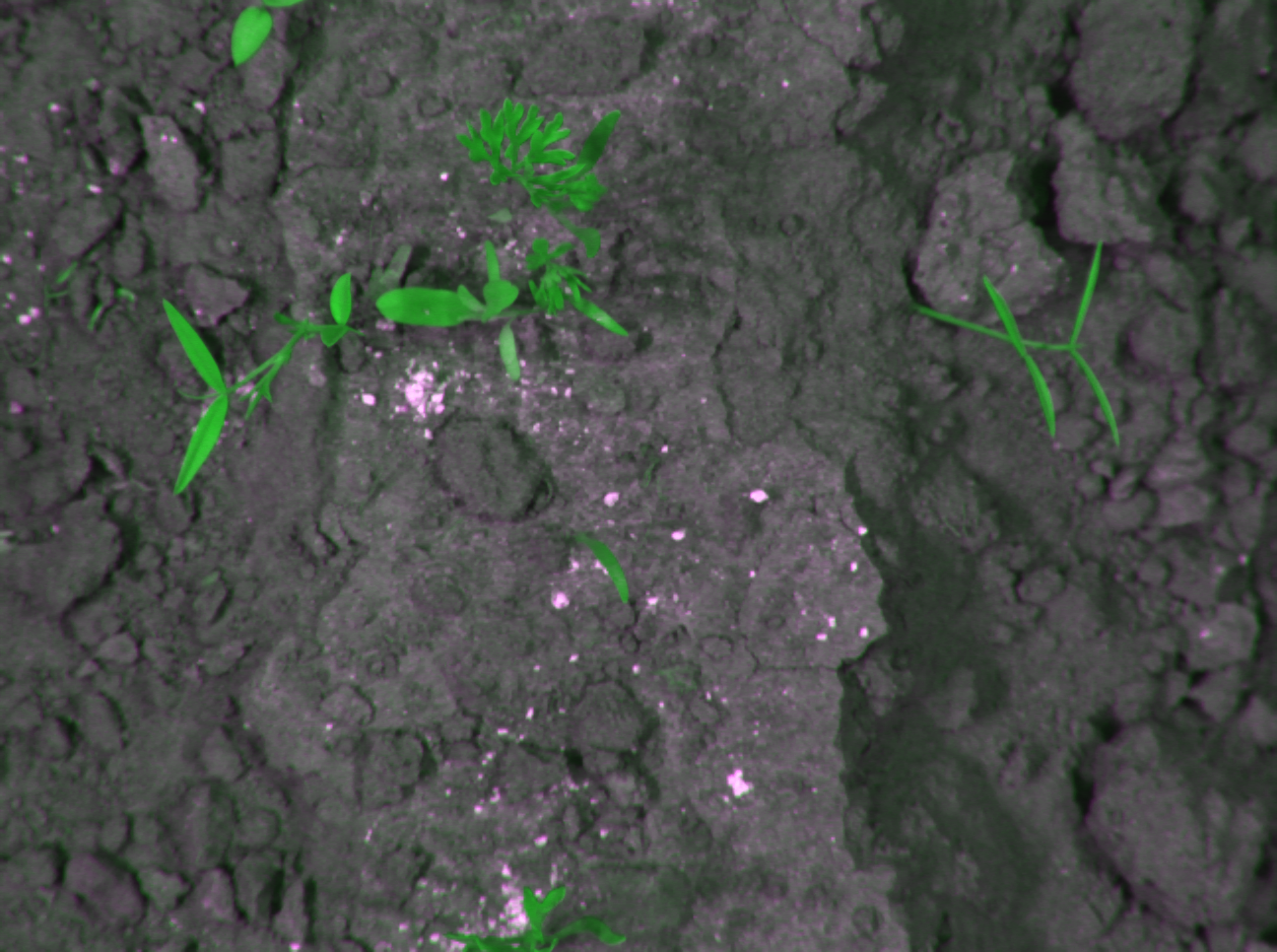}\hfill
    \includegraphics[width=.18\textwidth, height=0.16\textwidth]{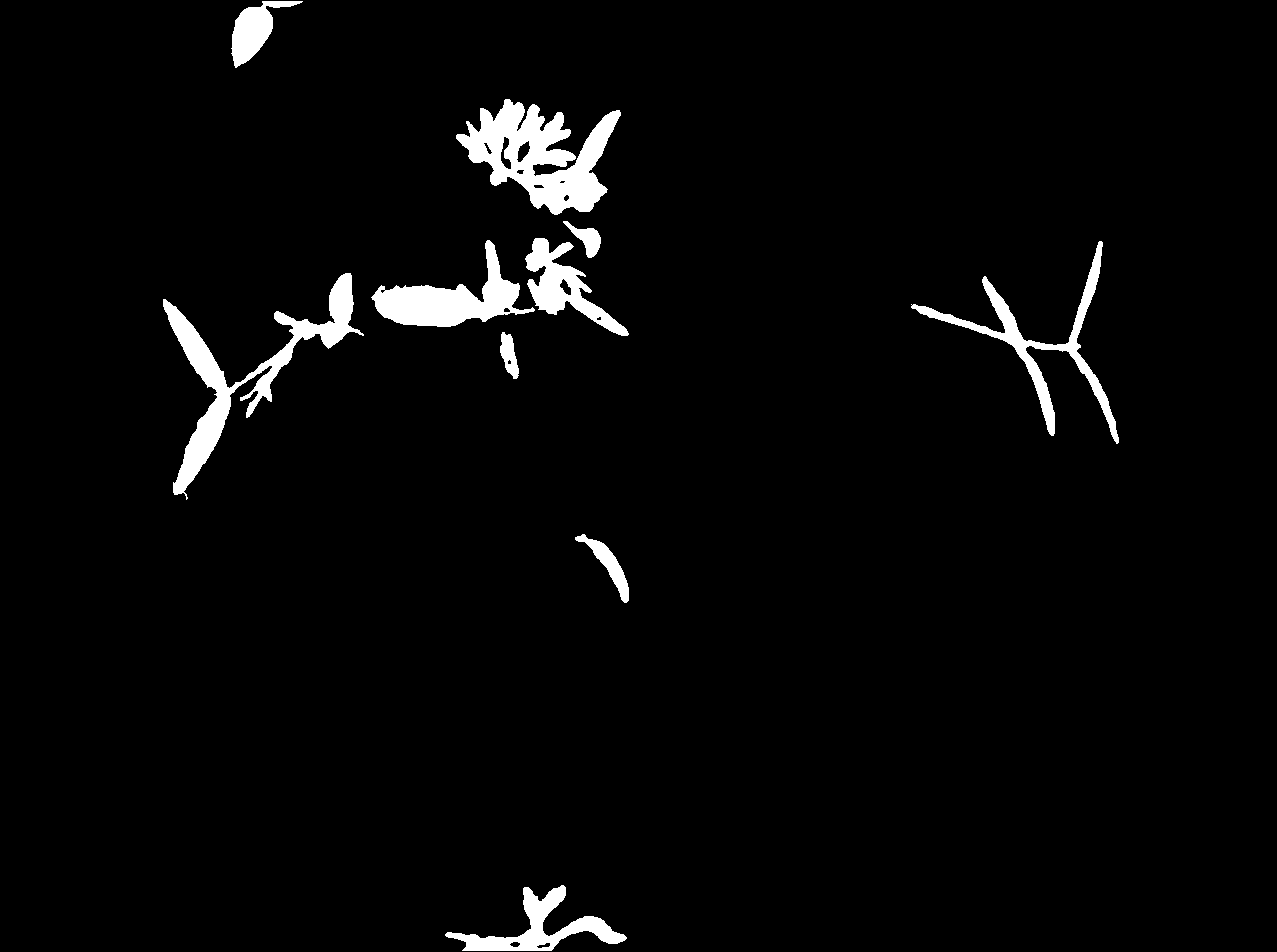}\hfill
    \includegraphics[width=.18\textwidth, height=0.16\textwidth]{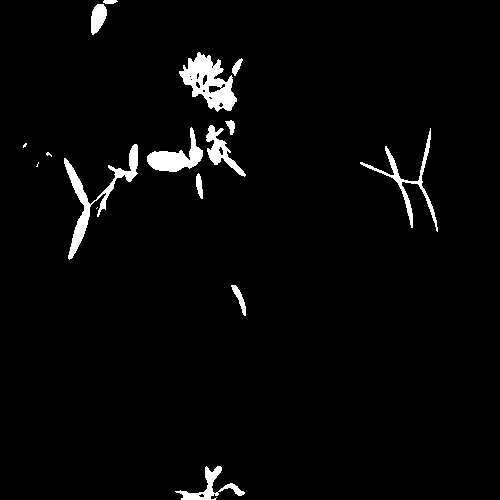}\hfill
    \includegraphics[width=.18\textwidth, height=0.16\textwidth]{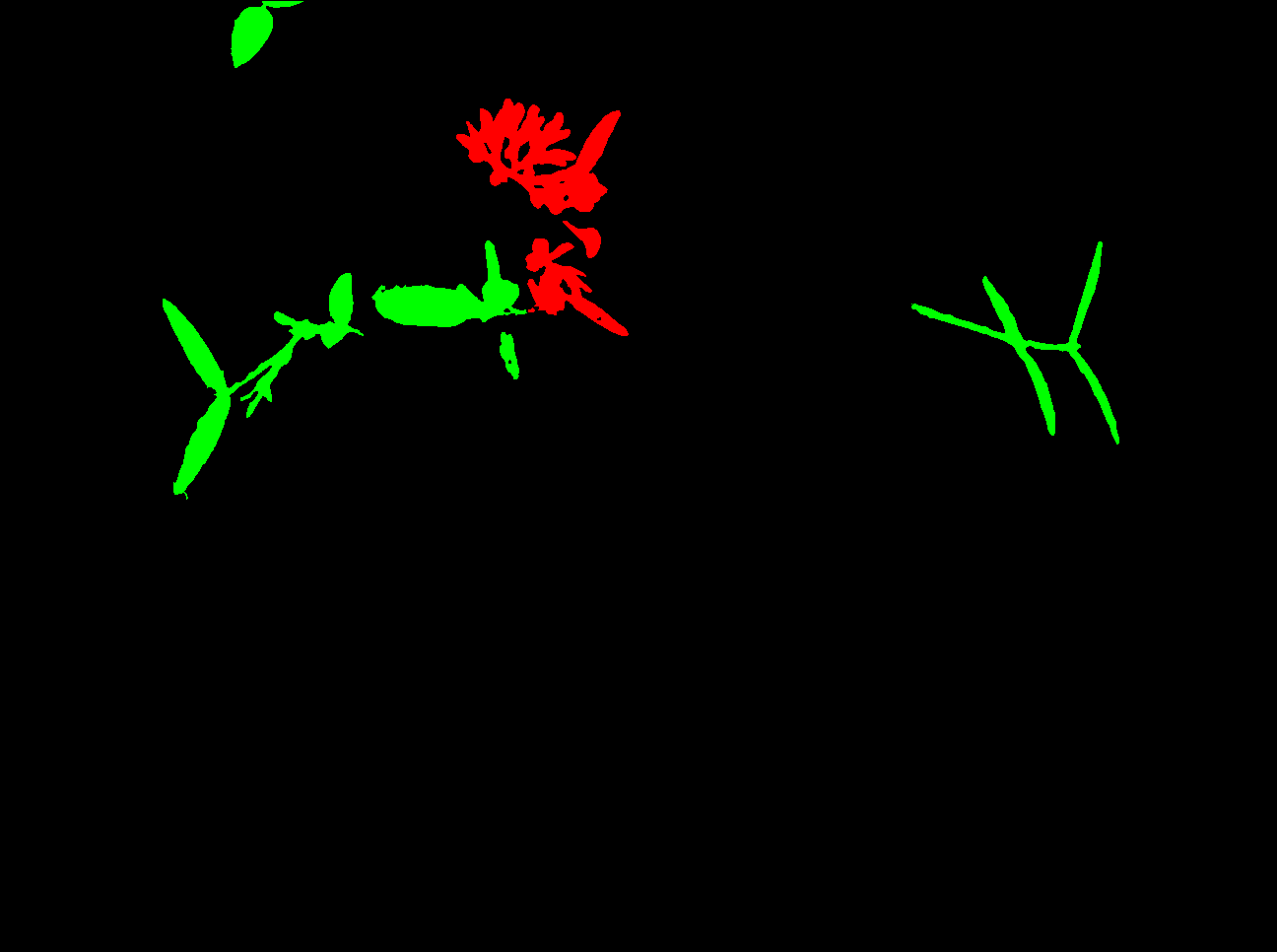}\hfill
    \includegraphics[width=.18\textwidth, height=0.16\textwidth]{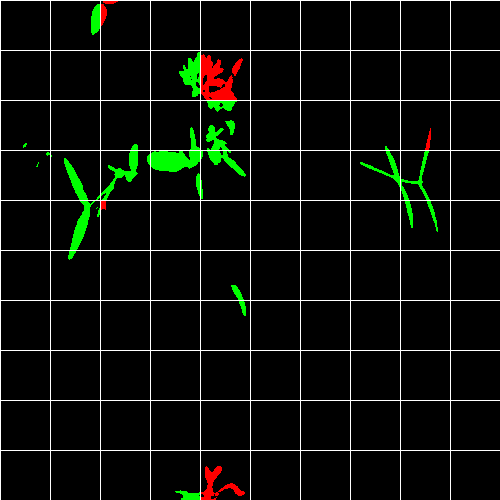}

    \vspace{0.35cm}
    \includegraphics[width=.18\textwidth, height=0.16\textwidth]{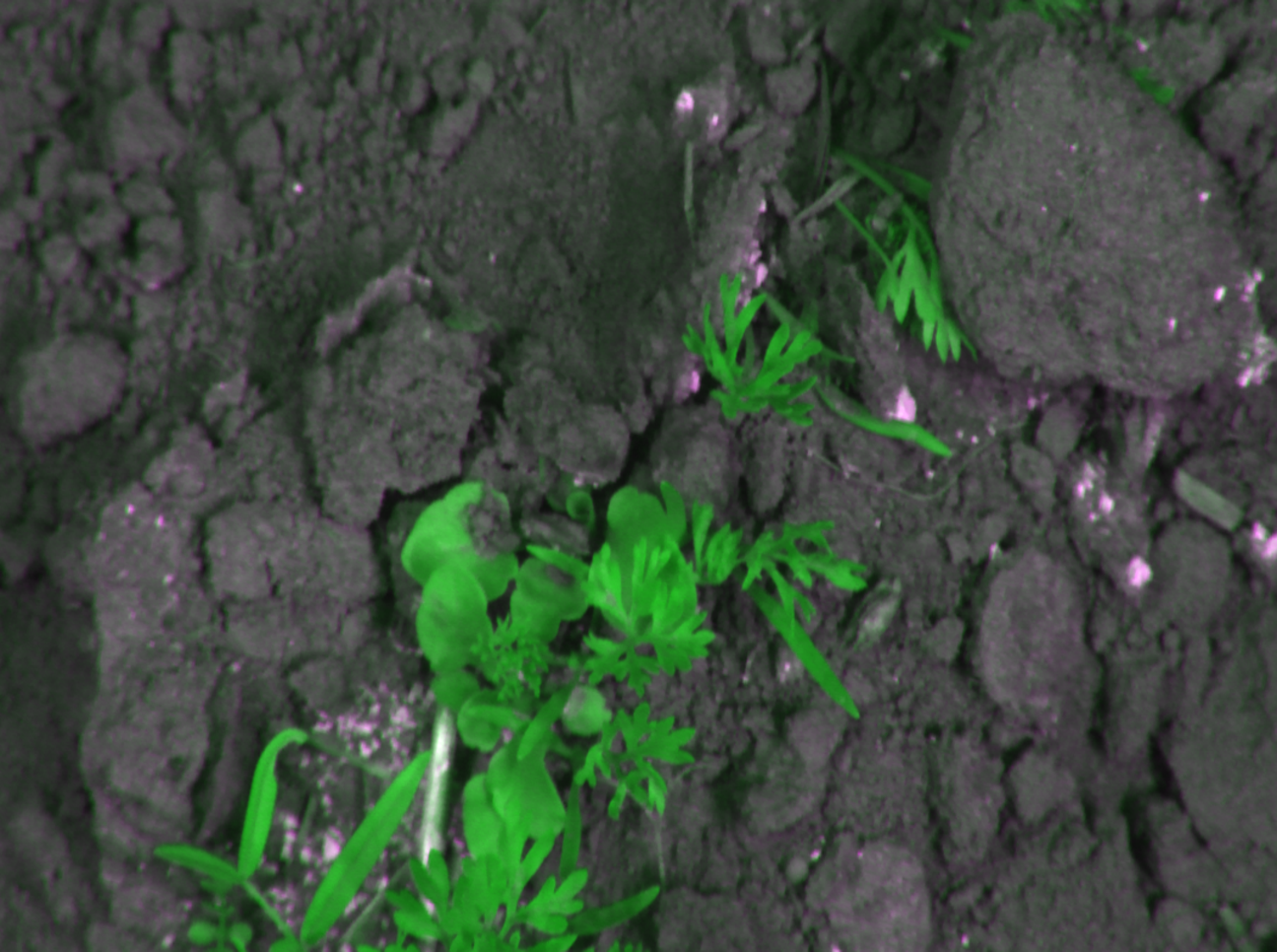}\hfill
    \includegraphics[width=.18\textwidth, height=0.16\textwidth]{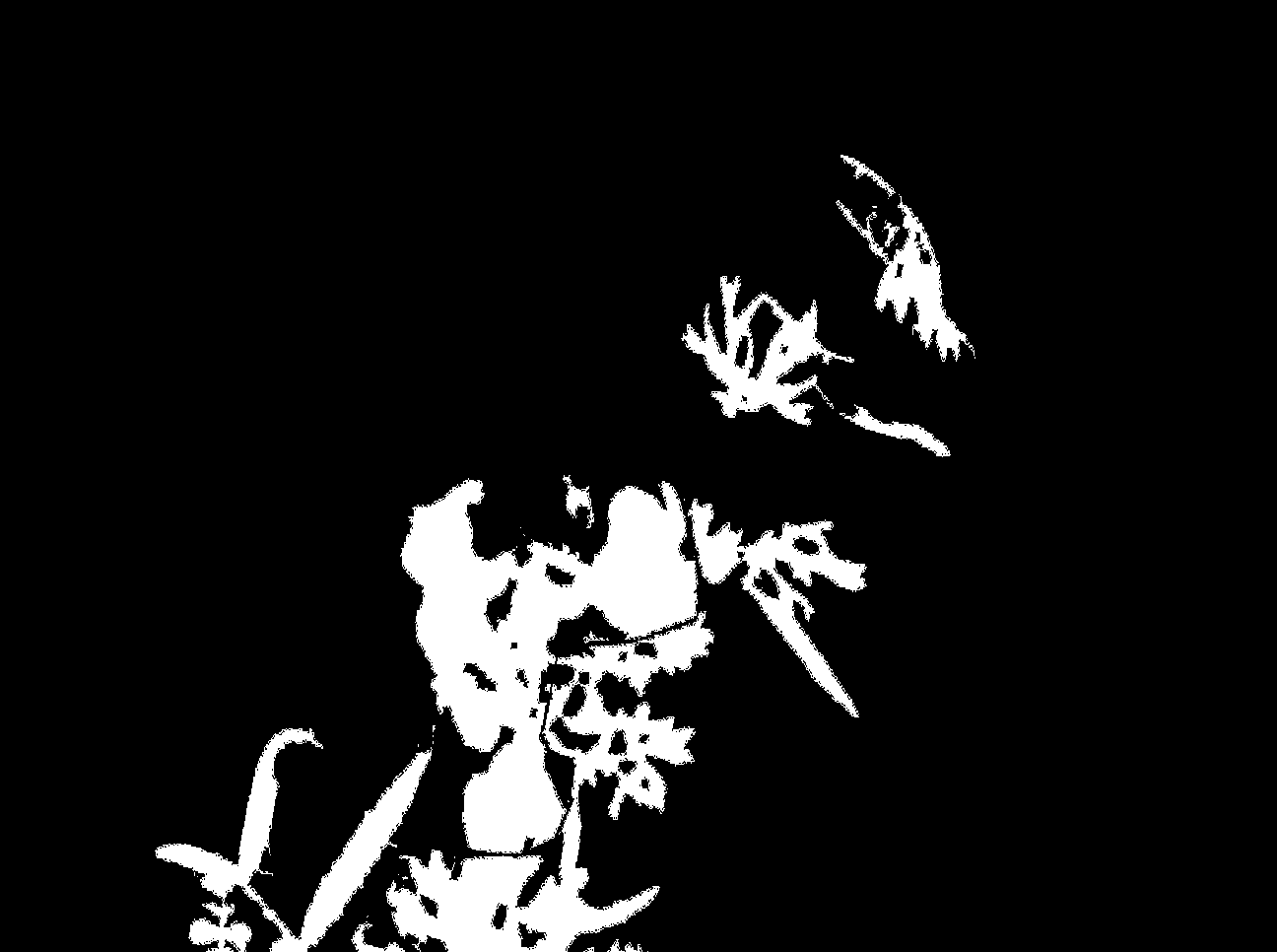}\hfill
    \includegraphics[width=.18\textwidth, height=0.16\textwidth]{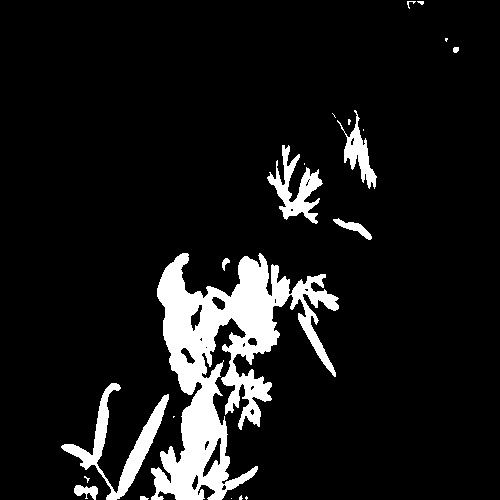}\hfill
    \includegraphics[width=.18\textwidth, height=0.16\textwidth]{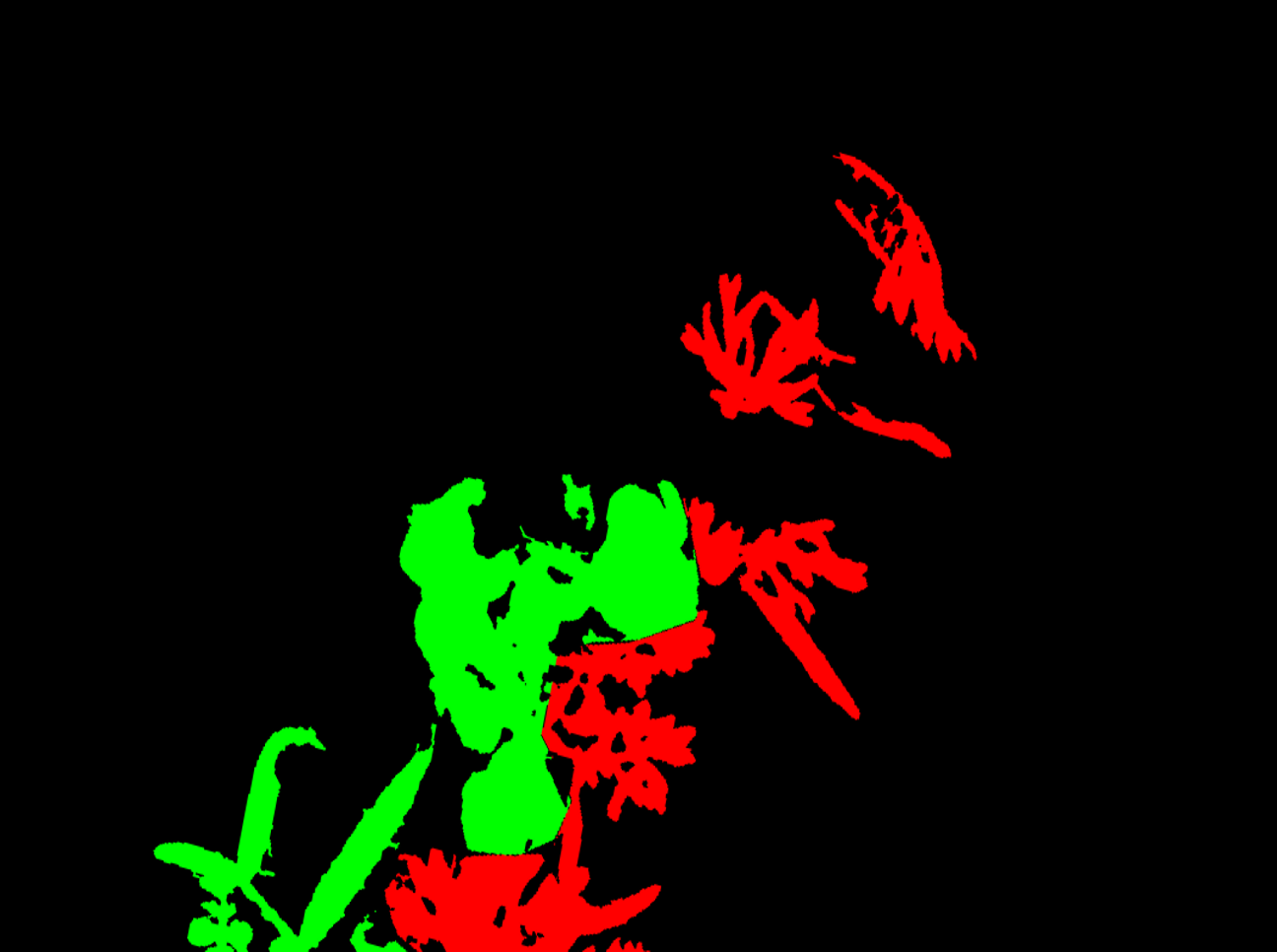}\hfill
    \includegraphics[width=.18\textwidth, height=0.16\textwidth]{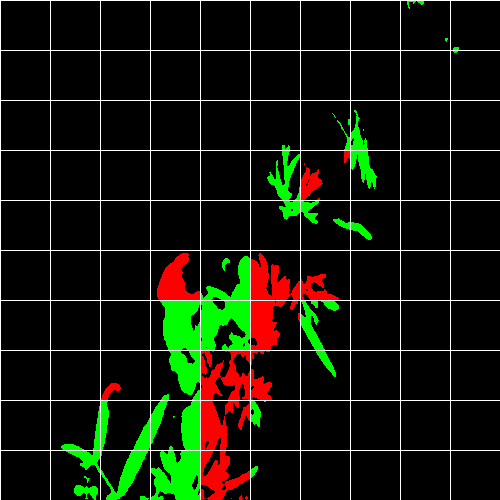}
    \caption{Result on the CWFID dataset: (left to right) (a) Input color image, (b) Ground truth vegetation mask, (c) Segmented vegetation mask, (d) Ground truth crop (red) and weed (green) pixels, (e) Predicted crop and weed pixels using fine-tuned ResNet50 as the classifier.}
    \label{fig:results_cwfid}
\end{figure*}

\begin{figure*}
    \includegraphics[width=.18\textwidth, height=0.17\textwidth]{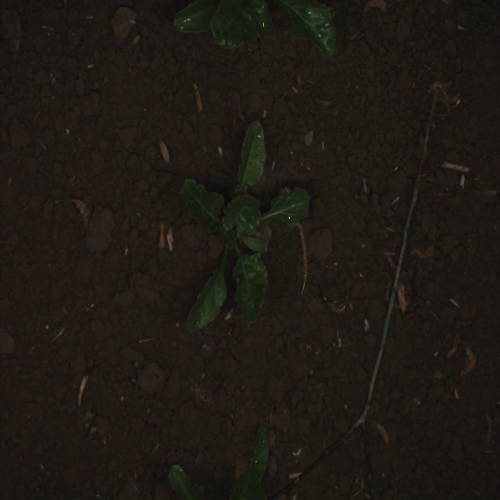}\hfill
    \includegraphics[width=.18\textwidth, height=0.17\textwidth]{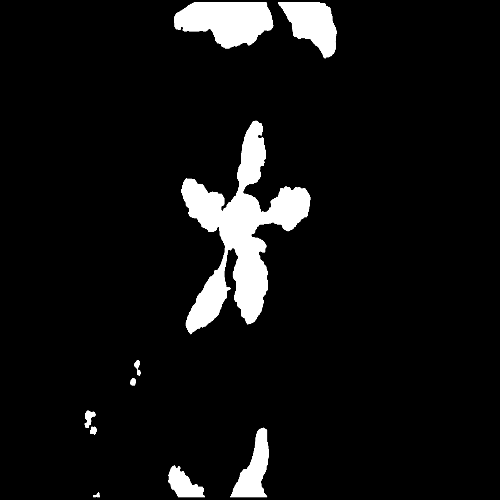}\hfill
    \includegraphics[width=.18\textwidth, height=0.17\textwidth]{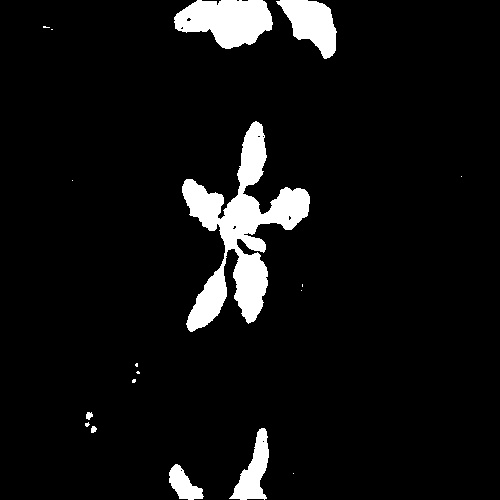}\hfill
    \includegraphics[width=.18\textwidth, height=0.17\textwidth]{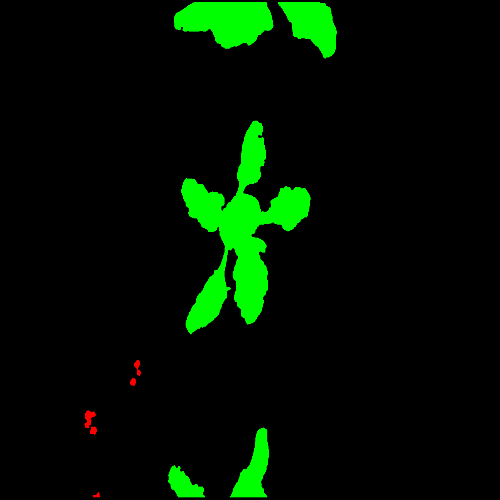}\hfill
    \includegraphics[width=.18\textwidth, height=0.17\textwidth]{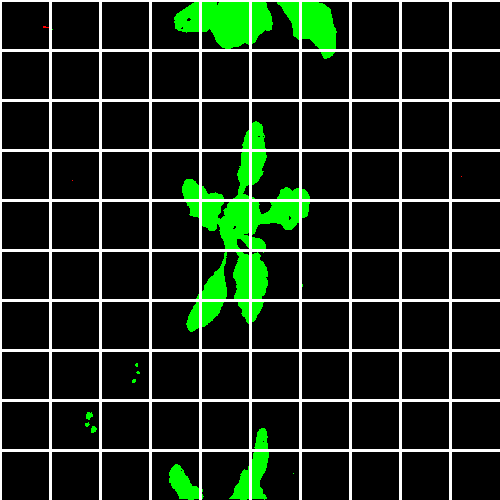}

    \vspace{0.35cm}
    \includegraphics[width=.18\textwidth, height=0.16\textwidth]{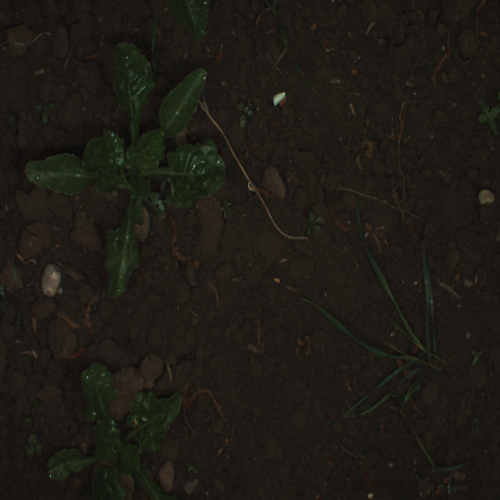}\hfill
    \includegraphics[width=.18\textwidth, height=0.16\textwidth]{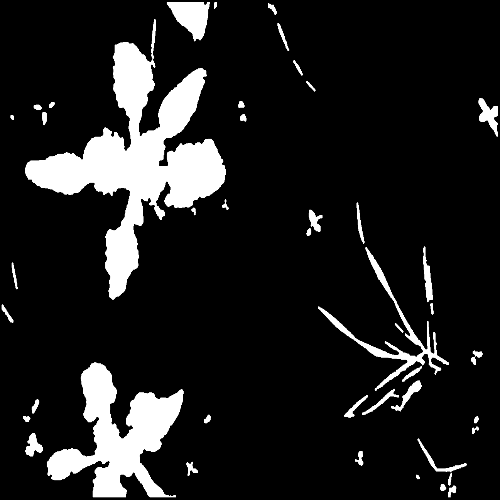}\hfill
    \includegraphics[width=.18\textwidth, height=0.16\textwidth]{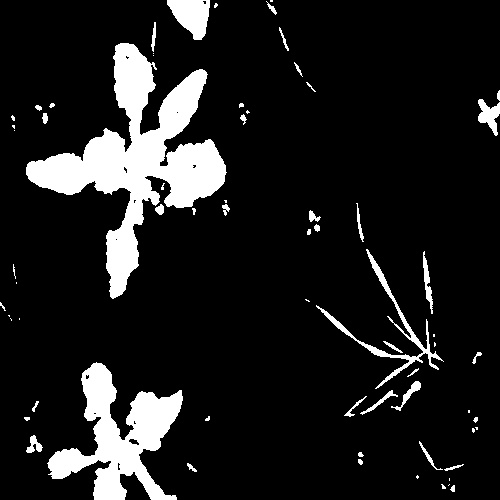}\hfill
    \includegraphics[width=.18\textwidth, height=0.16\textwidth]{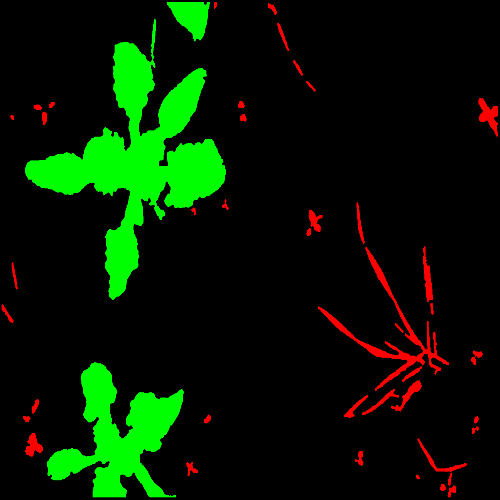}\hfill
    \includegraphics[width=.18\textwidth, height=0.16\textwidth]{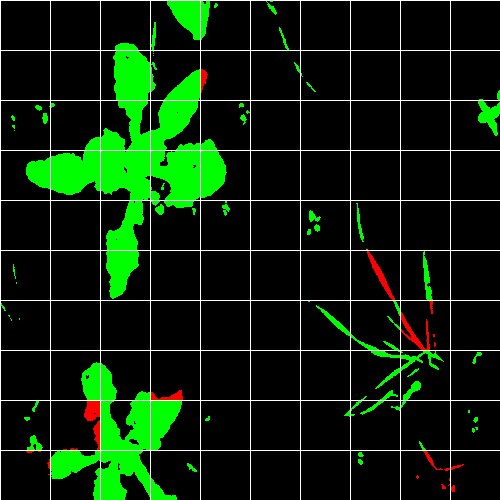}

    \caption{Result on the Sugar Beet Dataset : (left to right) (a) Input color image, (b) Ground truth vegetation mask, (c) Segmented vegetation mask, (d) Ground truth crop (red) and weed (green) pixels, (e) Predicted crop and weed pixels using fine-tuned ResNet50 as the classifier.}
    \label{fig:results_sb}
\end{figure*}

Despite the poor performance of the classical machine learning classifiers as a whole, the comparison was useful in determining the ideal tile size and validating weighted training of the networks. For image-based classififcation, ResNet50 was fine-tuned to classify $I_{tile}$ as crop or weed with the same tile size. The computed precision and recall values for weed/crop class are reported in Table \ref{resnet_comparison}. Weed precision is higher compared to feature-based classification with relatively higher recall value. On the other hand, the recall is lower for crop class but the precision is greater than 0.95 for weighted cross-entropy loss for both datasets. Besides, a recall of more than 0.91 is obtained for the weed class on both the datasets. This trend suggests that the network is likely to classify a region as weed-infested unless it is extremely confident that the region is free of any unwanted vegetation. This behaviour aligns with the objective to not overlook any weed-infested regions. It should be noted that given the larger number of crop tiles, this decision-making approach will correctly result in a large number of crop regions (with no weed plant coverage) not being treated with chemicals hence reducing their consumption significantly. Therefore, image-based classification using ResNet50 (with weighted cross entropy loss) is preferred over feature vector based classification. Figures \ref{fig:results_cwfid} and \ref{fig:results_sb} visualize the results for sample images from both the datasets.

%It may be noted that a recall of more than 0.9 for the weed class on both datasets signifies a very low false negative. As discussed earlier, recall is the preferred metric for evaluating the classifier performance because a high recall would ensure a low misclassification of weed-infested regions.  
 
\begin{table*}
    \centering
    \caption{Precision and recall values for classification using ResNet50 on the two datasets. (Class 0 : Crop, Class 1 : Weed)}. 
    \begin{tabular}{|c|c|c|c|c|c|c|c|c|c|c|c|c|}
        \hline
        Dataset & \multicolumn{6}{c|}{Cross-Entropy Loss} & \multicolumn{6}{c|}{Weighted Cross-Entropy Loss} \\
        \hline
        & P0 & R0 & F1 & P1 & R1 & F1 & P0 & R0 & F1 & P1 & R1 & F1\\
        \hline
        CWFID & 0.94 & 0.72 & 0.81 & 0.31 & 0.87 & 0.45 & 0.95 & 0.60 & 0.74 & 0.32 & 0.91 & 0.47 \\ 
        \hline
        Sugar Beets & 0.84 & 0.99 & 0.90 & 0.18 & 0.49 & 0.26 & 0.99 & 0.56 & 0.72 & 0.37 & 0.99 & 0.53 \\
        \hline
    \end{tabular}
    \label{resnet_comparison}
\end{table*}

\subsection{Weed Density estimation}

Once the weed-infested regions are identified, the cluster rate for each tile can be computed from the segmented vegetation pixels. A comparison of the estimated cluster rate for the weed-infested regions with the ground truth values is presented in Table \ref{tab:wd}. The results show that the weed density can be estimated reasonably across both the datasets. The loss of weed density pixels can be attributed to four reasons : (1) discarding tiles or regions with vegetation cover less than 10\% of the total area, (2) incorrect vegetation segmentation, (3) misclassification of weed-infested regions as crop plants and (4) presence of overlapping plants in a given tile. The first error source arises as the regions with low vegetation are ignored. The threshold, fixed at 10\% in this study, can be varied for different crop and weed plant requirements. The results reported for vegetation segmentation and weed distribution show that the proposed approach results in a mean absolute error of 5\% on CWFID and 1\% on Sugar Beets datasets. This indicates that the proposed method can handle the error arising due to above listed sources reasonably. Moreover, the RMSE of less than 8\% on two datasets of two different crop/weed species demonstrates that the proposed approach is scalable and can be adapted to any crop/weed species.  Once the weed plant distribution and density have been estimated, it is possible to make a decision about which regions should be selectively treated with agrochemicals.

\begin{table}[ht]
    \centering
    \caption{Accuracy of weed density estimation. MAE and RMSE are for a region of 2500 pixels.}
    \begin{tabular}{|c|c|c|c|}
        \hline
        Dataset & Mean Accuracy & MAE & RMSE \\
        & (\%) & (\%) & (\%) \\
        \hline
        CWFID & 75.24 & 5.02 & 7.85 \\
        \hline
        Sugar Beets & 82.13 & 1.62 & 3.06 \\
        \hline
    \end{tabular}
    \label{tab:wd}
\end{table}

\subsection{Comparison of pixel-wise dense predictions}

Although the proposed method's focus is not to predict an accurate pixel-wise weed/crop segmentation, the patch wise predictions can be used to generate the same. Hence, we compare the accuracy of the predicted ground coverage using the F1 score metric (Equation \ref{eqn:F1Score}).

%\begin{equation}
%    F1 = %\frac{\mbox{2*Precision*Recall}%}{\mbox{Precision + Recall}}
%    \label{eqn:f1}
%\end{equation}

The authors in  \cite{2, 3} proposed end-to-end segmentation networks for predicting dense crop/weed maps on Sugar Beets dataset. These methods report class-wise F1 scores where the maximum-minimum value for crop class is (0.9113, 0.9074), and for weed class is (0.8247, 0.7388). In comparison, our approach lags in terms of pixel-wise accuracy (maximum F1 value for crop class is 0.7153, and weed class is 0.3676). This can primarily be attributed to the reason that the end-to-end segmentation network can classify each pixel individually based on the features of its local area. However, in our approach, all the pixels belonging to a tile are classified as either crop or weed pixels regardless of the individual characteristics. In addition, we generate the vegetation segmentation, which contributes to error since few weed or crop pixels will be classified as background (soil) and vice-versa. On the other hand, segmentation networks have a single source of error as they segment and classify the pixels together.

However, for the purpose of selectively treating particular regions, the segmentation networks need to be augmented with an algorithm to select specific regions. If it is divided into regions such as square tiles, there is bound to be an overlap of weed and crop pixels for most of the tiles. The decision to treat a particular region will be taken from the dominant label for such tiles. Hence, the pixels which are correctly classified but are in the minority for a given tile do not influence the selective treatment. We argue that in the proposed approach, the focus is not on correctly identifying such pixels but correctly identifying the \textit{regions} to be treated (which can be robustly estimated as shown previously).  Besides, the volume of data required for the proposed method is significantly lower than that of an end-to-end segmentation network, enhancing generalizability and scalability. In addition, the proposed approach can be extended to any crop-weed combination as it eliminates the need to design hand-crafted features based on biological morphology and visual textures of the crop and weed. It may be noted that, to the best knowledge of the authors, there is no existing study on designing an end-to-end pixel-wise supervised CNN based segmentation approach on the CWFID dataset. One possible reason could be a limited number of images available in the CWFID dataset. However, encouraging results have been obtained on the CWFID dataset using the proposed tile-based semi-supervised approach. 

It may also be noted that the proposed tile-based system can cover the entire area of the original image by assigning a label to each tile. Hence, eventually, the total area being analyzed is the same, whether it is pixel-wise or tile-wise classification.

\section{Conclusion}
\label{conclusion}
Precision agriculture is described as a farmland management approach to maximize productivity and profits in a sustainable manner. Agrochemicals, such as weedicides, are an expensive input for farming in addition to being detrimental to the environment. Leveraging a computer vision system to identify regions for selective chemical treatment holds the potential to reduce their consumption significantly. In this paper, a semi-supervised approach to robustly estimate the weed density and distribution to aid precision agriculture is presented. The proposed approach relies only on color images as input. The first step is to generate a binary vegetation mask by removing all the background pixels. An unsupervised network is used to cluster the pixels into either background or vegetation. The second step is to overlay the mask on the input color image and divide it into smaller regions (square tiles of side 50 pixels). These smaller regions are then classified as weed or crop. In this work, the performance of two types of classifiers are studied: a) classifiers such as SVM, Gaussian Naive Bayes, Neural Network, and Random Forest which uses a pre-trained ResNet50 as a feature extractor and b) a fine-tuned ResNet50. The proposed approach is validated on two datasets consisting of different crop/weed species - Crop/Weed Field Image \cite{cwfid_dataset} and Sugar Beets \cite{sugar_dataset}. Weed infested regions are identified with a maximum recall of 0.99 and weed density in these regions is estimated with an accuracy of 82.13\%.

One of the primary objectives of our work is to reduce the dependency on extensively annotated datasets. The use of unsupervised segmentation and pre-trained ResNet50 in the proposed work eliminates the need for designing a hand-crafted features for weed identification. Compared to previous approaches, it is shown that it is possible to estimate both the weed distribution and density without training an end-to-end \textit{pixel-wise} segmentation network. Indeed, identification of weed-infected \textit{regions} could also aid in design of a robust site-specific weed management system.  The proposed pipeline is robust to varying plant species, overlapping plants, and images with poor contrast. This approach should help agricultural companies who are looking for low cost implementations as it requires very little training data and fine tuning. There is no need to invest in any extra sensors besides a regular RGB camera as long as there is a platform set up to collect top views of the plants. 

One of the limitations of our work is the iterative nature of generating vegetation masks. Future work should aim to reduce the average number of iterations required by the unsupervised network to generate the vegetation mask. This would improve the time needed to process a single color image on-board an autonomous robot. Another future direction of this work is to extend the two-stage detection and localization approach to medical imaging for identifying diseases or lesions. Such an approach can also be taken for identifying crop diseases to further expand the scope of precision agriculture.

\small{
    \bibliographystyle{unsrt}   
    \bibliography{ref}
}

\end{document}